\documentclass[12pt]{article}

\usepackage{xcolor}
\usepackage{scicite}

\usepackage{times}

\newenvironment{sciabstract}{%
\begin{quote} \bf}
{\end{quote}}

\usepackage{booktabs}
\usepackage{amsmath}
\usepackage{amssymb}
\usepackage{mathtools}
\usepackage{subcaption}
\usepackage{graphicx}
\usepackage[hidelinks]{hyperref}
\usepackage{xcolor}
\usepackage{multirow} 
\usepackage{fancyhdr}

\newcommand{\blue}[1]{#1}
\newcommand{\blu}[1]{#1}

\fancypagestyle{firstpage}
{
    \fancyhead[C]{\footnotesize{This preprint has not undergone any post-submission improvements or corrections.\\ The Version of Record of this article is published in npj Robotics, and is available online at \url{https://doi.org/10.1038/s44182-025-00047-y}}}
}

\title{Embodied Visuomotor Representation}

\author
{Levi Burner,$^{1\ast}$ Cornelia Ferm\"{u}ller,$^{2}$ Yiannis Aloimonos$^{2,3}$\vspace{10pt}\\
\normalsize{$^{1}$Department of Electrical and Computer Engineering, University of Maryland, College Park}\\
\normalsize{$^{2}$Institute for Advanced Computer Studies, University of Maryland, College Park}\\
\normalsize{$^{3}$Department of Computer Science, University of Maryland, College Park}\\
\normalsize{$^\ast$Corresponding author. Email: lburner@umd.edu (Levi Burner)}
}

\date{}

\begin{document} 


\baselineskip24pt


\maketitle 
\thispagestyle{firstpage}


\begin{sciabstract}
\begin{center}
{\large\textbf{Abstract}}
\end{center}
\blue{
Imagine sitting at your desk, looking at objects on it. You do not know their exact distances from your eye in meters, but you can immediately reach out and touch them. Instead of an externally defined unit, your sense of distance is tied to your action's embodiment. In contrast, conventional robotics relies on precise calibration to external units, with which vision and control processes communicate. We introduce {\it Embodied Visuomotor Representation}, a methodology for inferring distance in a unit implied by action. With it a robot without knowledge of its size, environmental scale, or strength can quickly learn to touch and clear obstacles within seconds of operation. Likewise, in simulation, an agent without knowledge of its mass or strength can successfully jump across a gap of unknown size after a few test oscillations. These behaviors mirror natural strategies observed in bees and gerbils, which also lack calibration in an external unit.
}

\end{sciabstract}


\newpage
\section*{Introduction}
\begin{figure}
  \centering
  \includegraphics[width=1.0\linewidth]{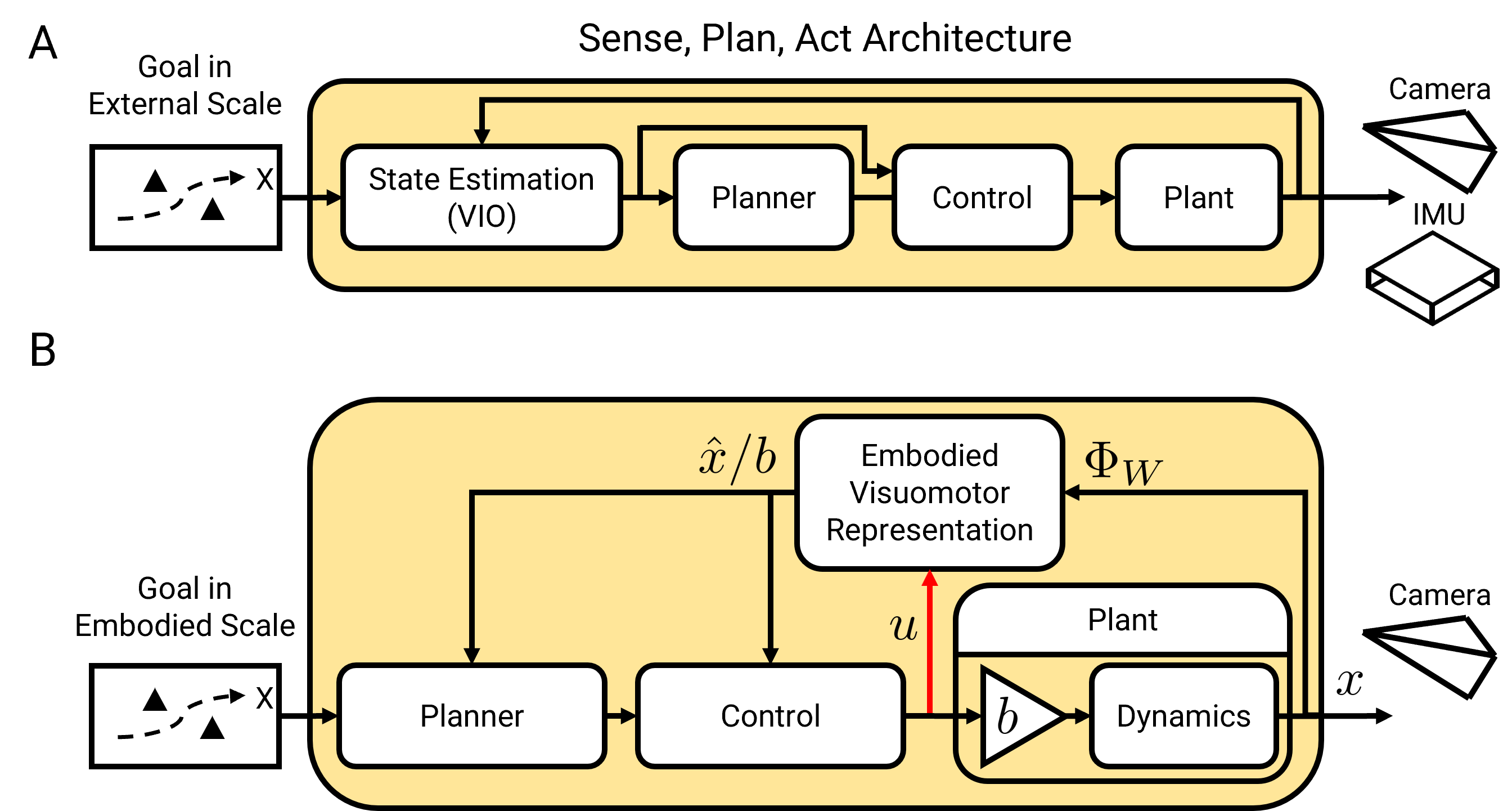}
  \caption{(A): The classic sense-plan-act architecture used in robotics assuming visual inertial odometry (VIO) is used for state estimation. Stability depends on calibrated sensors, such as an IMU, that provide accurate knowledge of the state in an external scale. (B): An architecture based on Embodied Visuomotor Representation. Compared to sense-plan-act, the embodied approach includes an additional internal feedback connection (red arrow) \blue{containing the control signal $u$. The units of $u$ are implied by the unknown gain $b$ and the dynamics going from control to the state in any scale, including the meter. Embodied Visuomotor Representation leverages position-to-scale, $\Phi_W$, obtained from vision and $u$ to determine a state estimate $\hat{x}/b$  in the embodied scale of $u$. Notably, the unknown $b$ cancels in the closed-loop system, enabling stable control without calibrated sensors. Direct methods such as Tau Theory, Direct Optical Flow Regulation, and Image Based Visual Servoing also avoid dependence on calibrated sensors by using purely visual cues (e.g., time-to-contact, optical flow, or tracked image features) for feedback. However, with few exceptions, such methods' stability depends on tuning the control law for the expected scene distances and system velocities.}
  }
  \label{fig:embodied}
\end{figure}

The \blue{predominant} autonomy frameworks in robotics rely on calibrated 3D sensors, \blue{predefined} models of the \blue{robot's} physical form, and structured representations of environmental interactions. \blue{These representations} allow vision and low-level control to be abstracted \blue{as separate processes that rely on} an external scale, such as the meter, to coordinate. For \blue{instance, it is common to use vision to construct a} metric map scaled to the meter. \blue{Subsequently, a planning algorithm uses this geometric representation to generate} a trajectory, \blue{also} scaled to the meter. \blue{Finally,} a pre-tuned low-level controller \blue{employs} feedback to follow the metric trajectory, mapping it to motor signals. This \blue{approach is known as} the sense\blue{-}plan\blue{-}act paradigm, \blue{and it originates from}  Marr's \blue{vision framework} \cite{marr1982}. Figure \ref{fig:embodied} \blue{illustrates} a block diagram of \blue{this} paradigm.

The sense-plan-act architecture allows separate teams of engineers and scientists to create equally separate vision and control algorithms tuned for particular tasks and mechanical configurations. Subsequently, alternative frameworks for vision\blue{,} such as Active Vision and Animate Vision emerged \cite{aloimonos1988active, bajcsy2018revisiting, BALLARD199157}\blue{,} primarily to address the limitations of the passive role of vision in the sense-plan-act cycle. However\blue{,} these frameworks consider control (action) at a high level, and so vision and control remain mostly separate fields and continue to be interfaced with external scale. This dependence leads to lengthy design-build-test cycles and, consequently, expensive systems that are available in only a few physical configurations and must be precisely manufactured. Further, it creates problems that are unobserved in biological systems. For example, it is well-known that Advanced Driver-Assistance Systems (ADAS) in modern cars, such as lane-keeping assist and collision warning, require millimeter accurate, per vehicle calibrations. Another example is the 1999 NASA Mars Climate Orbiter, whose software confused English and Metric units and crashed \cite{nasa1999mars}. Finally, several billion-dollar 3D Camera and LIDAR industries \blue{aim} to produce accurate 3D measurements calibrated to an external scale, and significant effort is dedicated to ensuring those calibrations do not degrade over time.

\blue{
This situation is striking because biological systems do not necessarily understand distance in an external scale like the meter. Further, we know from psychology that human and other mammalian perceptual systems do not provide metric depth and shape \cite{feldman1985four,koenderink1992surface,todd2004visual,cheong1998effects,ji2006noise}.  Instead, somehow, biological systems represent the world using only the signals due to their embodiment. Further, animals such as gerbils, dragonflies, and bumblebees exhibit visuomotor capabilities far in excess of robots while allowing for much broader variation in physical form and handling complex dynamics such as turbulence and muscular response \cite{ellard1984distance, mischiati2015internal, ravi2020bumblebees}. Similarly, humans can control augmentations to their embodiment, such as cars and airplanes, which also vary in physical form and do not maintain precise calibrations to an external scale.
}

\blue{
One explanation is that these systems do not rely on any precise sense of distance. In support, direct approaches such as Tau Theory \cite{lee1976theory}, Direct Optical Flow Regulation \cite{ruffier2005}, and classical Image Based Visual Servoing \cite{Chaumette2006RAMa} offer methods to control a visuomotor system using only cues directly available from the image such as time-to-contact, optical flow, and image features. However, systems strictly limited to direct methods cannot reason about distance, a capability essential for behaviors such as the clearing maneuvers of bees \cite{ravi2020bumblebees} and the jumping ability of gerbils \cite{ellard1984distance}. Moreover, direct methods either assume a predefined operating region for which the control system is tuned \cite{Chaumette2006RAMa, Chaumette2006RAMb, herisse2012, de2022accommodating} or constrain achievable behaviors to those possible with a restricted class of feedback laws \cite{Chaumette1999, Corke2001}. In contrast, systems with knowledge of distance can overcome these limitations while still using direct (image space) control objectives.}

\blue{
Thus, a key question arises: How can a robot, without a source of external scale, use its vision and motor system to estimate a meaningful sense of distance?}
We refer to methods that capture this phenomenon as {\it Embodied Visuomotor Representation } and propose that \blue{its} realization could drive a paradigm shift that makes robots \blue{easier to produce, more affordable, and widely accessible.}

Psychologists assume humans have such representations without proposing a precise and general method for obtaining them \cite{Glenberg_1997}. Biologists argue that simple animals such as bees \blue{can} develop them \cite{ravi2020bumblebees}. Computer scientists suggest that \blue{computations drive intelligence}, and thus representations, as physical as the body itself \cite{brock2024intelligencecomputation}.
In particular, there is evidence that the visuomotor capabilities of animals can be attributed to well-tuned internal models closely coupled with perceptual representations. \blue{In addition to reasoning about distances, an} important function of these models is prediction, which allows an animal to anticipate upcoming events, compensate for the significant time delays in their feedback loops, and, when coupled with inverse models, generate feedforward signals for other parts of the body \cite{Wolpert2019}.

However, being able to explain a phenomenon is not the same as being able to implement it. \blue{To address this}, we propose Embodied Visuomotor Representation, a concrete framework that can estimate embodied, action-based distances using an architecture that couples vision and control as a single algorithm. This results in robots that can quickly learn to control their visuomotor systems without any pre-calibration to a unit of distance. In many ways, our method parallels a psychological theory of memory as embodied action due to Glenberg (who presents the idea of embodied grasping of items on a desk). Glenberg argues that the core benefit of embodied representation is that they ``do not need to be mapped onto the world to become meaningful because they arise from the world'' \cite{Glenberg_1997}. Similarly, our Embodied Visuomotor Representations do not need to be pre-initalized with an external scale because they can be estimated by the robot online without sacrificing stability.

Embodied Visuomotor Representation \blue{achieves} this by exchanging distance on an external scale for the unknown but embodied distance units of the acceleration \blue{affected} by the motor system. As a consequence, powerful self-tuning, self-calibration, or adaptive properties emerge. At the mathematical level, the approach is similar to the Internal Model Principle, a control theoretic framework that \blue{underlies all learning-based control methods, either explicitly or implicitly} \cite{huang2018}. \blue{Unlike the sense-plan-act approach, which operates sequentially, the Internal Model Principle introduces a critical topological distinction: an internal feedback loop that leverages embodied motor signals to predict sensory feedback representations}. This same internal feedback loop, shown in Figure \ref{fig:embodied} enables Embodied Visuomotor Representation to integrate vision and control seamlessly, without relying on external scale, or sacrificing close-loop stability.

In what follows, we develop \blue{a general} mathematical form for equality constrained Embodied Visuomotor Representation and demonstrate it on a series of experiments where uncalibrated robots gain the ability to touch, clear obstacles, and jump gaps after a few seconds of operation \blue{by using natural strategies observed} in bees flying through openings and gerbils jumping gaps.
\blue{For these applications, we develop specific relations between control inputs, acceleration, and observed visual features, such as lines and planes, that can be used by an uncalibrated robot to estimate state in an embodied unit. This state is used by control schemes that are guaranteed to perform a task correctly.}
The results suggest that Embodied Visuomotor Representation is a paradigm shift that will allow all visuomotor systems, and in particular robots, to automatically learn to control themselves and engage in lifelong adaption without relying on pre-configured 3D sensors, controllers, or 3D models that are calibrated to an external scale.

\section*{Results}
First, the mathematical methodology for Embodied Visuomotor Representation will be detailed in the equality constrained case, where the results of vision and control can be set equal to each other. Next, examples of applying Embodied Visuomotor Representation to the control of a double integrator and a multi-input system with actuator dynamics are developed. Finally, algorithms that use Embodied Visuomotor Representation to allow uncalibrated robots to learn to touch, clear obstacles, and jump gaps are presented. Detailed calculations behind these applications are provided in the Methods section.

\subsection*{Equality Constrained Embodied Visuomotor Representation}
Consider a point $X_w \in \mathbb{R}^{\blue{4}}$ \blue{in homogenous coordinates corresponding to the 3D location of a point in the} world coordinate frame. \blue{It is transformed by the extrinsics matrix} $T_{cw} \in SE(3)$ and projected to \blue{the homogenous} pixel coordinates $p_c \in \mathbb{R}^3$ in the image \blue{according to an invertible intrinsics matrix $K \in \mathbb{R}^{3\times3}$ that encodes the focal length and center pixel}. The resulting well known relationship is

\begin{equation}\label{eqn:projection}
p_c = \frac{1}{X^z_c} \begin{bmatrix}K & 0\end{bmatrix} T_{cw} X_w.
\end{equation}

\blue{Where $X_c^z$ is the third element of the product $T_{cw} X_w$.} Without loss of generality, we assume that $K = I$.

Next, consider the warp function as known in computer vision. It has the property that
\begin{equation}\label{eqn:warp}
p_c(t) = W(t, t_0, p_c(t_0)).
\end{equation}

That is, given the initial position of a world point in an image at time $t_0$, the warp function returns the position of \blue{the \blu{same}} world point in the image at another time. \blu{Note that warp functions are sometimes assumed to be linear, and this can be a good approximation when the region of interest is on a flat plane \cite{baker2004lucas}. However, in general, the warp is nonlinear because its estimation amounts to the integration of optical flow or tracking visual features such as points, lines, or patches.}

Similarly \blu{to the warp function,} the ``flow map'' from control theory, $\Phi_{f,u}$, is defined as returning the solution to a system defined by $f$ with input $u$ as follows:

\begin{equation}\label{eqn:ode}
\begin{split}
\dot{x} &= f(x, u), \quad x(t_0) \in \mathbb{R}^n\\
x(t) &= \Phi_{f, u}(t, t_0, x(t_0))
\end{split}
\end{equation}

Comparing \eqref{eqn:warp} to \eqref{eqn:ode} reveals that $W$ and $\Phi$ serve the same purpose. $W$ is the solution map giving trajectories in image space, \blue{whose derivative is also called optical-flow,} and $\Phi$ is the solution map for the system state\blue{, whose time derivative is driven by the system dynamics. This similarity in purpose leads to an equality constraint.}

Due to $W$'s close relationship to position $X$ through \eqref{eqn:projection}, many visual representations allow computing the position of the camera up to a characteristic scale such as the size of an object under fixation, the initial distance to a visual feature, or the baseline between two stereo cameras. \blu{We call the position to scale $\Phi_W$ and assume it can be estimated. In particular, the prominent families of computer vision algorithms, such as homography estimators, structure from motion, SLAM, and visual odometry result in such a $\Phi_W$.}

\blu{$\Phi_W$ can be related to position by a multiplicative scalar, that is the characteristic scale of the visual representation, which we call $d$.} \blu{I}f the first three elements of the state are the position of a point in the camera's frame, i.e., $x = \begin{bmatrix} X_c^T & *\end{bmatrix}^T$, \blu{then $\Phi_W$, $d$, and $\Phi_{f, u}$ are related by,}

\begin{equation}\label{eqn:controlvisioncouple}
\begin{split}
X_c(t) = \Phi_W(t, t_0, p_c(t_0)) d = \Phi_{f, u}^{1,2,3}(t, t_0, x(t_0)),
\end{split}
\end{equation}
where $p_c(t_0) = x^{1,2}(t_0) / x^3(t_0)$ and superscripts (i.e., $\Phi_{f, u}^{1,2,3}$) are used to denote the individual components of a vector.  \blu{Note that} this formulation holds for loose and tight couplings of vision and control, where loose coupling means the vision algorithm functions independently without the knowledge of action.

The problem of visuomotor control thus becomes finding specific forms of $\Phi_{f,u}$ and $W$ \blue{that possess desirable} properties. In particular, we \blue{seek} a transformation that \blue{reformulates} the problem \blue{into} a form \blue{akin to those} used in linear system theory,
\blue{enabling} us to take advantage of the properties of linear systems. Suppose the \blu{camera is translating but not rotating. Then the world frame's axes can be assumed to align with the camera frame's axis.} In practice, this \blue{assumption holds for durations ranging from} a few to several seconds \blue{when} an inexpensive Inertial Measurement Unit is mounted in the same frame as the camera sensor \blu{and is used to compensate the image so that it can be considered as coming from a camera with fixed orientation}. \blue{Consequently}, the rotation between the camera frame $c$ and the world frame $w$ can be neglected \blue{as} the frames can be assumed to coincide up to a translation \blu{and so} $X$ can be used in the place of $X_c$.

Then Newton's second law of motion can be applied in the \blue{fixed} camera frame (within the time interval that the system can be considered rotation invariant). Let that time interval be  $[t_0, t_0+T]$ where $T \neq 0$. Then we obtain a linear system that relates the position, velocity, and acceleration components of the state $X$ and $\Phi_W$,

\begin{equation}\label{eqn:visuomotormodel}
\begin{split}
\Phi_W(t, t_0) d &= X(t_0) + (t-t_0)\dot{X}(t_0) + \int_{t_0}^{t} \int_{t_0}^{\sigma} f_{mech}(X(\sigma_2), \dot{X}(\sigma_2), x_{act}(\sigma_2), u(\sigma_2), \theta) d\sigma_2 d\sigma.
\end{split}
\end{equation}

Here $f_{mech}$ encapsulates the mechanical dynamics due to forces exerted by actuators, $\theta$ is a set of mechanical parameters (arm lengths, mass, etc.) that can be considered constant, $\dot{X}$ is the time derivative of $X$, and $x_{act}$ is the state of the actuator dynamics. We encapsulate the actuator dynamics as a separate system

\begin{equation}\label{eqn:visuomotormodelact}
\dot{x}_{act}(t) = f_{act}(x_{act}(t), u(t), \theta_{act}),
\end{equation}
where $\theta_{act}$ are the constant parameters of the actuators.

Consider this visuomotor representation as known by an embodied agent without calibration to an external scale. Then $X$ is the robot's 3D position relative to the point of interest in an unknown unit, $d$ is the characteristic scale of vision in \blue{the unknown} unit, $\Phi_W$ is the unitless estimate of the position due to vision,  $\theta$ and $\theta_{act}$ are the constant parameters of the mechanical system and the actuators in unknown units. \blue{Finally,} $x_{act}$ and $u$ are the state of the actuators in unknown units.

The only quantities known in general by an uncalibrated embodied agent system are then $\Phi_W$ and $u$. But, because $\Phi_W$ is unitless, if the embodiment attempts to estimate  $X$, $d$, $\theta$, and $\theta_{act}$ so that \eqref{eqn:visuomotormodel} and \eqref{eqn:visuomotormodelact} hold at all times, the units of the estimated quantities must be implicitly determined by $u$. For example, it can be seen that the distance units of $X$ are determined by the distance units defined by the acceleration, that is, a distance unit over seconds squared, whose magnitude is determined from the embodied signal $u$ transformed into acceleration by the system\blue{'}s $f_{act}$ and $f_{mech}$. In other words, \blue{the magnitude of} $X$, $\dot{X}$, and $d$ \blue{is determined by the  magnitude of the \blue{the action} $u$ and thus their units are implied by $u$.}

Consequently, the general path for employing Embodied Visuomotor Representation involves \blue{the following steps:}

\begin{itemize}
    \item \blue{Define a model} that represents the physical \blue{dynamics} of the agent. Both classical and learning-based models can be \blue{used}.
    \item Ensure all quantities in the model \blue{arising from physical processes are unit-free or up to scale}. For example, if gravity is to be modeled, it can be assumed the effect is non-zero, but the magnitude must not be assumed.
    \item Choose a visual representation $\Phi_W$ where the characteristic scale is related to a quantity of interest. For instance, \blue{in a grasping task}, it may be advantageous to choose $\Phi_W$ so that $d$ corresponds to the \blue{object's size}. Similarly, $d$ \blue{might be} the initial distance to a tracked feature, line, or object \blue{in a navigation task}.
    \item \blue{Define a constraint between the unknowns} $d$, $X$, $\dot{X}$, $\theta$, $\theta_{act}$, and $x_{act}$ \blue{that enables a unique and task-relevant representation to be estimated}. In what follows, we give two specific formulations.
    \item Construct an estimator for the chosen representation of $d$, $X$, $\dot{X}$, $\theta$, $\theta_{act}$, and $x_{act}$ \blue{and incorporate these estimates into the task execution.}
\end{itemize}

\subsection*{Closed Loop Control with Embodied Visuomotor Representation}
In what follows, we give two specific examples of constructing an estimator and using the resulting state for closed-loop control. In the first example, distance is estimated in units of action for a single input, single output system. In the second example, the multiple input case is considered. \blue{In that case,} distance is most directly available in \blue{multiples} of the characteristic scale of vision but can easily be converted to the \blue{units} of action of any of the inputs. In both cases, stable closed-loop control is a consequence of the embodied representation.

For each example, we begin by constructing a model, a sliding window estimator, and a closed-loop controller which will have guaranteed stability under mild assumptions. The approach is a form of indirect-adaptive control because, first, a forward model is identified using our framework. Then, a controller is synthesized from that model \cite{astrom2008adaptive}.

The simplest system that can be represented in the framework of Embodied Visuomotor Representation \blue{is a robot moving along the optical axis while fixating on a nearby object. The dynamics are assumed to be that of a double integrator and the image provides the position up to the characteristic scale} of vision. This system was also considered in \cite{TTCDist}, however, the exposition was limited. The system is given by

\begin{align}\label{eqn:doubleint}
    &\dot{x} = \underbrace{\begin{bmatrix} 0 & 1 \\ 0 & 0\end{bmatrix}}_{A}x + \begin{bmatrix} 0 \\ b\end{bmatrix} u\\
    &\blu{\Phi_W(t, t_0) d = x_1(t)}.
\end{align}

\blu{Here, the observation (output) is $\Phi_W$, and it is linearly related to the state $x_1$ through the unknown $d$.} $b$ is the unknown gain between control effort and acceleration in an external scale, such as meters per second squared. $b$ implicitly depends on both the strength of the embodied agent's actuators and the mass of the embodiment itself. Dimensionality analysis reveals that $b$ is simply the unknown conversion factor between the embodied scale implied by $u$ and the external scale. \blue{Finally,} the system is observable as long as $d\neq0$. \blu{That is, the visual representation's characteristic scale, such as the initial distance to a visual feature, size of a patch, or the baseline between two stereo cameras, is non-zero.}

A sliding window estimator that considers $\Phi_W$, and $u$ to be known over the interval $[t_0, t_0+T]$, $T \neq 0$ results in the problem

\begin{equation}\label{eqn:externaldoubleintest}
\begin{split}
\min_{d, x(t_0), b} \int_{t_0}^{t_0+T} \left(
\Phi_W(t, t_0) d - x_{\blue{1}}(t_0) - (t - t_0)\blue{x_2}(t_0) - b \int_{t_0}^t \int_{t_0}^\sigma u(\sigma_2) d\sigma_2 d\sigma \right)^2 dt,
\end{split}
\end{equation}
which cannot be solved uniquely \blue{without additional constraints because \blu{a trivial solution can be realized by allowing all optimized variables to equal zero.}} However, dividing by $b$\blue{, the unknown conversion factor between external scale and embodied scale,} reveals a problem that can be solved as long as acceleration is non-zero for some period during the interval $[t_0, t_0 + T]$

\begin{equation}\label{eqn:embodieddoubleintest}
\begin{split}
\min_{\frac{\left[d, x(t_0)\right]}{b}} \int_{t_0}^{t_0+T} \left(
\Phi_W(t, t_0) \frac{d}{b} - \frac{x_{\blue{1}}(t_0)}{b} - (t - t_0)\frac{\blue{x_2}(t_0)}{b} - \int_{t_0}^t \int_{t_0}^\sigma u(\sigma_2) d\sigma_2 d\sigma \right)^2 dt.
\end{split}
\end{equation}

Since $b$ is simply a conversion factor between units, we see that the lumped quantities $d/b$, $x\blue{_1}(t_0)/b$, and $\blue{x_2}(t_0)/b$ are traditional state estimates but in the embodied units of $u$. Thus, we see that an embodied agent can naturally estimate state in an embodied scale by comparing what is observed with the accelerations effected through action.

\blue{
In practical applications, $\Phi_W$ will be noisy because it is estimated using an image made out of discrete pixels. Even if this noise is zero mean, Equation \eqref{eqn:embodieddoubleintest} results in a biased estimator because it uses $\Phi_W$ as an independent variable. An unbiased estimator can be realized by dividing by $d/b$, resulting in the problem,
}

\blue{
\begin{equation}\label{eqn:embodieddoubleintestunbiased}
\begin{split}
\min_{\frac{\left[b, x(t_0)\right]}{d}} \int_{t_0}^{t_0+T} \left(
\Phi_W(t, t_0) - \frac{x_{\blue{1}}(t_0)}{d} - (t - t_0)\frac{\blue{x_2}(t_0)}{d} - \frac{b}{d}\int_{t_0}^t \int_{t_0}^\sigma u(\sigma_2) d\sigma_2 d\sigma \right)^2 dt.
\end{split}
\end{equation}
}
\blue{In the new problem, $\Phi_W$ is the dependent variable \blu{ because it is no longer multiplied with one of the optimized variables. Then, because the problem is linear least squares, if $\Phi_W$} is corrupted with zero mean noise, the parameter estimates will remain unbiased. On the other hand, the estimated parameters have changed. The state $x$ is now estimated in multiples of $d$ instead of the desired multiples of $b$. Further, only the reciprocal of the $d/b$ is estimated.
These estimates can be transformed back to the desired quantities with division, which will result in some bias since division does not commute with expectation. However, unlike when using Equation \eqref{eqn:embodieddoubleintest}, techniques that increase the accuracy of estimates, such as taking more measurements, will increase the accuracy of the transformed estimates. A detailed explanation is provided in the Supplemental Material.
}

\blue{The solution to either formulation} is unique as long as the acceleration, $u$, is non-zero for a finite period within the sliding window.
\blue{A detailed proof is given in the Supplemental Material.}

Then if the control law $u = K (x/b)$ is applied, the closed loop system dynamics become

\begin{equation}\label{eqn:bcancelsdoubleint}
    \dot{x} = Ax - \begin{bmatrix} 0 \\ b\end{bmatrix} K \frac{x}{b} = Ax - \begin{bmatrix} 0 \\ 1\end{bmatrix}Kx.
\end{equation}

\blu{Because the unknown gain $b$ cancels out in the rightmost expression, we can say that the closed loop behavior of the system is invariant to the value of $b$. That is, while $b$ is not known individually, and is only a term in the known ratios $x/b$ and $d/b$, it will not affect the systems behavior. Thus, the control gains $K$ can be chosen as if $b=1$ and $x$ itself is known. This is despite the fact that $b$ appears individually in the dynamics model. Similarly, the characteristic scale of vision, $d$, does not affect the closed loop behavior despite remaining unknown.}

Further, consider that \blue{once} $d/b$ is known, $\Phi_W(t) d/b$ provides a direct estimate of $x_1(t)$ in the embodied unit. Thus, in theory, it is sufficient to solve \eqref{eqn:embodieddoubleintest} \blue{or \eqref{eqn:embodieddoubleintestunbiased}} once, and subsequently treat control of \eqref{eqn:doubleint} as a traditional output feedback problem. \blue{Finally, since the closed loop system is linear, time-invariant, controllable, and has known parameters, feedback gains $K$ always exist and can be chosen to ensure global closed-loop stability. In particular, since this system corresponds to a PD controller applied to a double integrator, the stabilizing gains can be easily chosen.}

\blue{Next, the previous example is extended to the case that the robot does not know its actuator dynamics. Typically, actuator dynamics produce a ``lagging'' response to control inputs. For example, they could take the form of an unknown delay or low-pass filter between the control input and its effect on acceleration. Additionally, the full 3D motion is considered.}

Suppose the dynamics are fully linear\blue{,} the robot can move in three dimensions\blue{, there is an unknown, stable linear system between control inputs and the 3D acceleration, and a visual process estimates the position of the robot up to the characteristic scale of vision.} Then\blue{,} the \blue{dynamics} can be expressed as

\begin{equation}
\begin{split}
&\dot{x} = \frac{d}{dt}
\begin{bmatrix}
X\\
\dot{X}\\
x_{act}\\
\end{bmatrix}
=
\begin{bmatrix}
\dot{X}\\
B u + C_{act} x_{act}\\
A_{act} x_{act} + B_{act}u
\end{bmatrix}\\
&\blu{\Phi_W(t, t_0) d  = X}
\end{split}
\end{equation}

Estimating the unknown parameters $B$, $A_{act}$, $B_{act}$, and $C_{act}$ simplifies to satisfying the following equality constraint

\begin{equation}\label{eqn:generalizedphiconst}
\begin{split}
&\Phi_W(t, t_0) d = X(t_0) + t \dot{X}(t_0)\\
&+ \int_{t_0}^t \int_{t_0}^{\sigma} B u(\sigma_2) + C_{act} \left[ \Phi_{A_{act}}(\sigma_2, t_0) x_{act}(t_0)  + \int_{t_0}^{\sigma_2} \Phi_{A_{act}}(\tau, t_0) B_{act} u(\tau) d\tau\right] d\sigma_2 d\sigma
\end{split}
\end{equation}

\blue{Suppose} that the actuator dynamics expressed inside the double integral are BIBO stable\blue{. Then,} the impulse responses relating the inputs of the actuator's system to this term go to zero exponentially fast. \blue{Consequently, it is sufficient in many practical applications to approximate these dynamics as a finite length convolution with the input.}
Then, we get the following equality constraint where $G$ is a matrix value signal whose $i$,$j$'th entry is to be convolved with the $j$'th \blue{element of $u$} to get its effect on the $i$'th \blue{state}.

\begin{equation}
\begin{split}
\Phi_W(t, t_0) d &= X(t_0) + (t-t_0) \dot{X}(t_0) + \int_{t_0}^t \int_{t_0}^{\sigma} (G*u)(\sigma_2)d\sigma_2 d\sigma.
\end{split}
\end{equation}

\blue{As with the previous example considering a double integrator}, any problem based on this constraint \blue{has no unique solution because every term has an unknown multiplier}. However, unlike the double integrator, we cannot simply divide by $b$ to get a problem with a unique solution. Instead, we can divide by the scalar $d$ to get a linear constraint \blue{that will be uniquely satisfied} given sufficient excitation \blue{by $u$}. \blue{A detailed proof is given in the Supplemental Material.} Consider the sliding window estimator again\blue{. The} full problem to be solved is

\begin{equation}\label{eqn:slidingadaptiveproblem}
\begin{split}
    \min_{\frac{[X(t_0), \dot{X}(t_0), G]}{d}} 
    \int_{t_0}^{t_0+T} \Bigg[&\Phi(t, t_0) -\frac{X(t_0)}{d} -(t-t_0) \frac{\dot{X}(t_0)}{d} 
    - \left(\frac{G}{d} * \int_{t_0}^{\boldsymbol{\cdot}} \int_{t_0}^{\sigma} u (\sigma) d\sigma_2 d\sigma  d\sigma\right)(t)\Bigg]^2 dt,
\end{split}
\end{equation}
where the fact that the convolution with $G$ can be brought out of the double integral has been used, and $\boldsymbol{\cdot}$ at the top of the double integrator is the placeholder for the variable that will be convolved over.

In this case, the position of the agent is recovered \blue{in multiples of the} characteristic scale of vision due to division by $d$. However, \blue{the} units of distance can still be recovered in the embodied units of acceleration \blue{by dividing} all estimated parameters by the gain of one of the actuator's impulse responses at a particular frequency. In particular, if the DC gain of the actuator dynamics is non-zero, we can consider that

\begin{equation}\label{eqn:dcgainscale}
\left(\frac{X(t_0)}{d}\right) / \left(\frac{\int_0^{\infty} g_{ij}(t) dt}{d}\right)
= \frac{X(t_0)}{\int_0^{\infty} g_{ij}(t) dt},
\end{equation}
Where $g_{ij}$ is the $i$, $j$ entry of $G$ \blue{and $\int_0^{\infty} g_{ij}(t) dt$ is its scalar valued DC gain}. Thus, distance in units of action is still available. \blue{Further,} the distance in the units of action is different for each input, and the conversion factor between all the embodied scales is known to the agent\blue{. So,} it is free to switch between units as may be convenient. 

If the actuator dynamics are minimum phase, and thus an inverse impulse response $G^{-1}$ exists, then since $G/d$ is known, $d G^{-1}$ can be determined. It is then \blue{straightforward} to synthesize a reference tracking controller. \blue{Consider the control scheme given by}

\begin{equation}
\begin{split}
e &\coloneqq \begin{bmatrix} X_{ref} \\ \dot{X}_{ref} \end{bmatrix} -\begin{bmatrix} X/d \\ \dot{X}/d \end{bmatrix}\\
u &\coloneqq d G^{-1} * Ke.
\end{split}
\end{equation}

\blue{Then the} closed-loop dynamics are

\begin{equation}\label{eqn:dynamicscontrolcharacteristic}
\begin{split}
\ddot{X} &= G*u = G*(dG^{-1} * Ke) = K \left(d\begin{bmatrix} X_{ref} \\ \dot{X}_{ref} \end{bmatrix} -\begin{bmatrix} X \\ \dot{X} \end{bmatrix}\right),
\end{split}
\end{equation}
where the fact that the systems $G$ and $dG^{-1}$ cancel (by definition) except for the scale $d$ has been used.

The resulting closed loop's stability does not depend on the magnitude of the embodied scale or any external scale. Instead, the reference point is in multiples of vision\blue{'s} characteristic scale, which could be the size of a tracked object, the baseline between a stereo pair, or the initial distance to an object. As before\blue{,} it is straightforward to convert to the embodied units of action because $d$ can be replaced \blue{with any actuators DC gain,} $\int_0^{\infty} g_{ij}(t) dt$\blue{. In that} case\blue{,} the closed loop dynamics become

\begin{equation}\label{eqn:dynamicscontrolembodied}
\begin{split}
\ddot{X} &= K \left(\int_0^{\infty} g_{ij}(t) dt \begin{bmatrix} X_{ref} \\ \dot{X}_{ref} \end{bmatrix} -\begin{bmatrix} X \\ \dot{X} \end{bmatrix}\right).
\end{split}
\end{equation}

\blue{As with \eqref{eqn:bcancelsdoubleint}, the systems in \eqref{eqn:dynamicscontrolcharacteristic} and \eqref{eqn:dynamicscontrolembodied} are linear, time-invariant, controllable, and all parameters are known. Thus, gains $K$ always exist and can be selected to ensure closed loop stability \cite{Hespanha2018}}.

In practice, the inverse of the actuator dynamics should not be used in the control law because the cancellation of dynamics typically results in a control law with poor performance and robustness. Regardless, we consider the inverse dynamics above so that the final close loop control law clearly illustrates the units being employed by Embodied Visuomotor Representation. However, in general, it is now straightforward to design traditional control laws using the identified parameters. In particular, the jumping experiments avoid using inverse dynamics.

We now turn to three basic robotic capabilities: touching, clearing, and jumping, which can be accomplished by uncalibrated robots that use Embodied Visuomotor Representation.

\subsection*{Uncalibrated Touching}

Consider an uncalibrated robot with a monocular camera that must touch a target object in front of it. The contact will occur at a non-zero speed because the robot does not know its body size and thus cannot come to a stop just as it reaches the target. Further, the robot also does not know the strength of its actuators, the size of the target, or the size of anything else in the world. \blue{In what follows, we carefully apply the five steps for using Embodied Visuomotor Representation to design an algorithm for this problem. Figure \ref{fig:touching} outlines the resulting uncalibrated touching procedure visually}.

\blue{Starting with the first step, we assume a double integrator system \blue{as in \eqref{eqn:doubleint}}, that is} the robot accelerates along \blue{the optical} axis according to $\blue{\ddot{x}_1} = bu$ where $\blue{\ddot{x}_1}$ is acceleration in meters per second squared and $u$ is a three-dimensional control input. \blue{Upon inspection, we see that the only quantity defined by the world, $b$, does not have a predetermined unit. Thus, the second step is complete.}

\blue{To complete the third step, assume the robot is facing a planar target. The reciprocal of the target's width in normalized pixels within the visual field can then be used as $\Phi_W$. That is, $\Phi_W = Z / d$, where $\Phi_W$'s characteristic scale is the unknown width of the target, $d$. It should be noted that this simple visual representation is valid only for flat objects that are coplanar with the camera, making it and following experiment pedagogical in nature. However, in general, \eqref{eqn:controlvisioncouple} demonstrates that Embodied Visuomotor Representation is agnostic to the particular visual representation or algorithm used so long as it provides position up to a characteristic scale.
}

\blue{The fourth step, which involves establishing a constraint between vision and control that includes the unknowns, is accomplished by recalling Equation \eqref{eqn:doubleint} and dividing through by $b$. This results in the unknowns $x(t_0)/b$ and $d/b$, which correspond to the initial conditions, $x(t_0)$, and the characteristic scale of vision, $d$, in the embodied units of $u$. Once this is done, the fifth step can be applied: estimating the unknowns using Equation \eqref{eqn:embodieddoubleintestunbiased}.}

\blue{
The unique estimate of $d/b$ obtained from this formulation allows the distance between the robot and the target to be estimated at all times via the relation $\hat{x}_1/b = \Phi_W(t)(d/b)$. This can be combined with the target's position in the image to recover the target's 3D position. Closed-loop control can then use this distance to center the robot on and approach the target along a direction in the camera frame at a constant speed, because in the closed loop, the remaining unknown, $b$, cancels out, as shown in Equation \eqref{eqn:bcancelsdoubleint}.
}

\blue{To ensure safe contact without damaging the robot, the designer must specify a safe contact speed, $v_s$, in embodied units. This limitation mirrors biological systems, where animals often create safe environments for their young to prevent injury, ensuring that the maximum achievable speed does not cause harm. It is important to note that the robot cannot stop just before touching the target, as it does not know how where its body is  relative to the camera.}

\blue{
In practice, the robot’s designer or a higher-level cognitive process can set $v_s$ based on multiples of the target's dimensions or by considering multiples of the maximum force (in embodied units) that the robot should experience upon collision. An example of this calculation is provided in the Supplemental Material. Specifically, if the designer knows the size of the target in meters and a safe speed in meters per second, they can convert the safe speed to multiples of the target’s size per second, which the robot can subsequently interpret in embodied units.
}

\blue{Subsequently,} contact with the object can be detected either with a touch sensor or by using the visual position estimate $\Phi_W d$ to detect that the robot has stopped getting closer to the target. Thus, the robot can approach a target at a constant safe speed in a given direction in the camera field and touch it, regardless of the embodiment and environment's specifications.

\blue{It should be noted that it was assumed that the open-loop oscillations would not cause the robot to come in contact with the obstacles. However, this is not strictly necessary. Equation \eqref{eqn:embodieddoubleintestunbiased} has a unique solution given any acceleration of arbitrarily small duration. Thus, in theory, it is always possible to estimate the unknowns and switch to closed loop control prior to reaching the target.}

\begin{figure}
  \centering
  \includegraphics[width=1.0\linewidth]{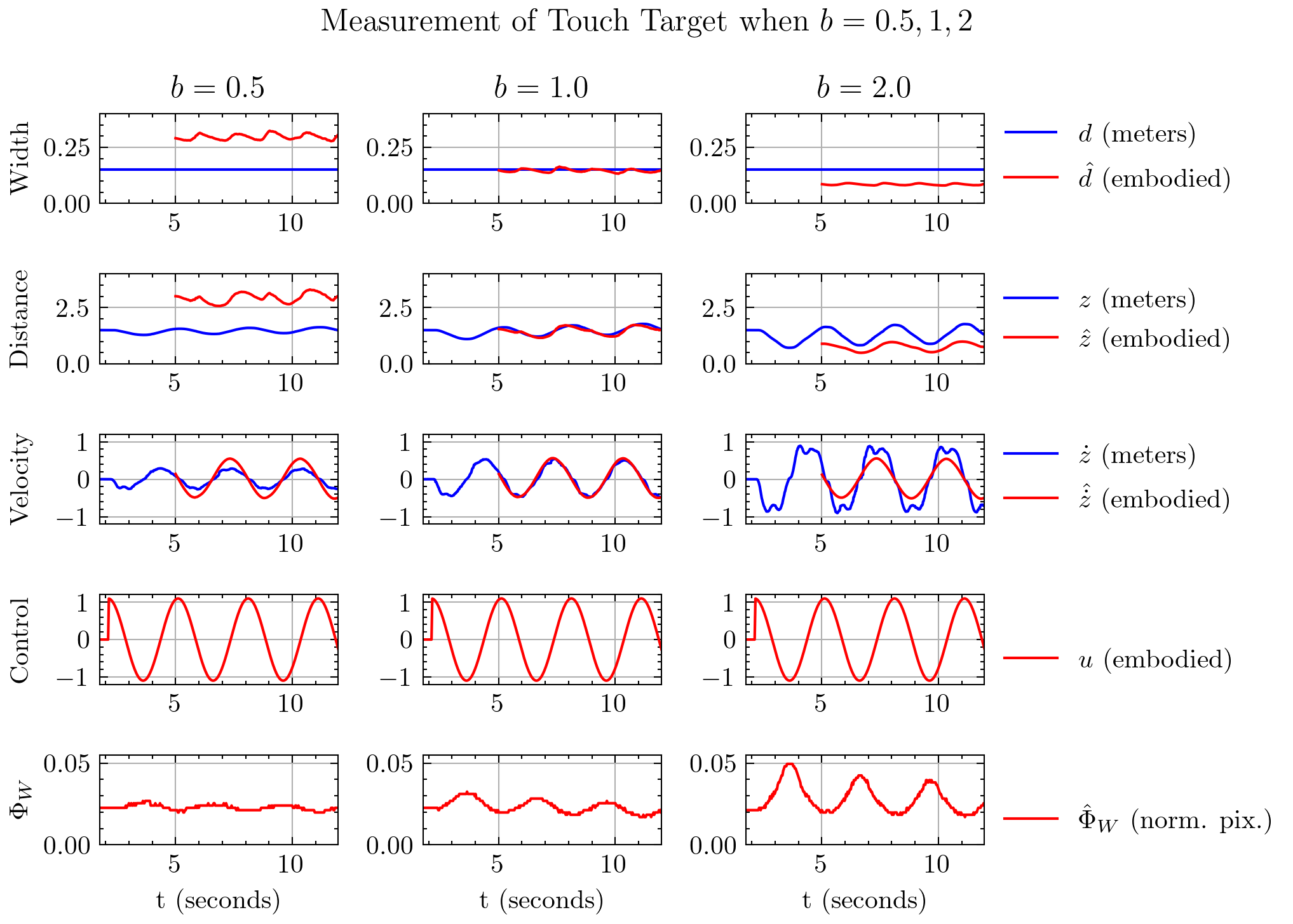}
  \caption{\blue{Experimental data and estimated signals from the target measurement phase of uncalibrated touching for three different values of the control gain $b$. The first, second, and third rows of plots show the estimated target width, distance from the target, and velocity as estimated by the sliding window formulation in Equation \eqref{eqn:bcancelsdoubleint}. The bottom two rows show the inputs to the sliding window formulation, that is, the control signal $u$ and the width of the target in the visual field in normalized pixel coordinates $\Phi_W$. As expected, when $b=$ 1.0, the embodied estimates are closely aligned with the ground truth distances measured in meters. However, when $b=$ 0.5 or $b=$ 2.0, the embodied estimates are twice as much and half, respectively, of the ground truth value in meters.}}
  \label{fig:measuretarget}
\end{figure}

Figure \ref{fig:touching} \blue{visually} outlines the \blue{uncalibrated touching procedure}. Movie 1 details the experimental procedure for both uncalibrated touching and clearing. \blue{Figure \ref{fig:measuretarget} shows the estimated width, distance, and velocity over time during the measurement phase of uncalibrated touching for three different control input gains, $b =$ 0.5,  1, and 2. Since the conversion from embodied units to meters is determined by $b$, the embodied estimate is expected to align with the ground truth only when $b=$ 1. Similarly, Figure \ref{fig:approachtarget} shows estimated distances and control inputs as the robot approaches and touches the target while using the same three ground truth values for $b$. The approach velocity was fixed to the same embodied value for each trial, meaning when $b=$ 0.5, the robot takes twice as long to reach the target compared to $b=$ 1. Whereas when $b=$ 2, it takes half the time. Alternatively, the methods discussed above for specifying a safe contact speed allow the robot's designer to ensure a consistent approach speed regardless of $b$.}

\begin{figure}
  \centering
  \includegraphics[width=0.5\linewidth]{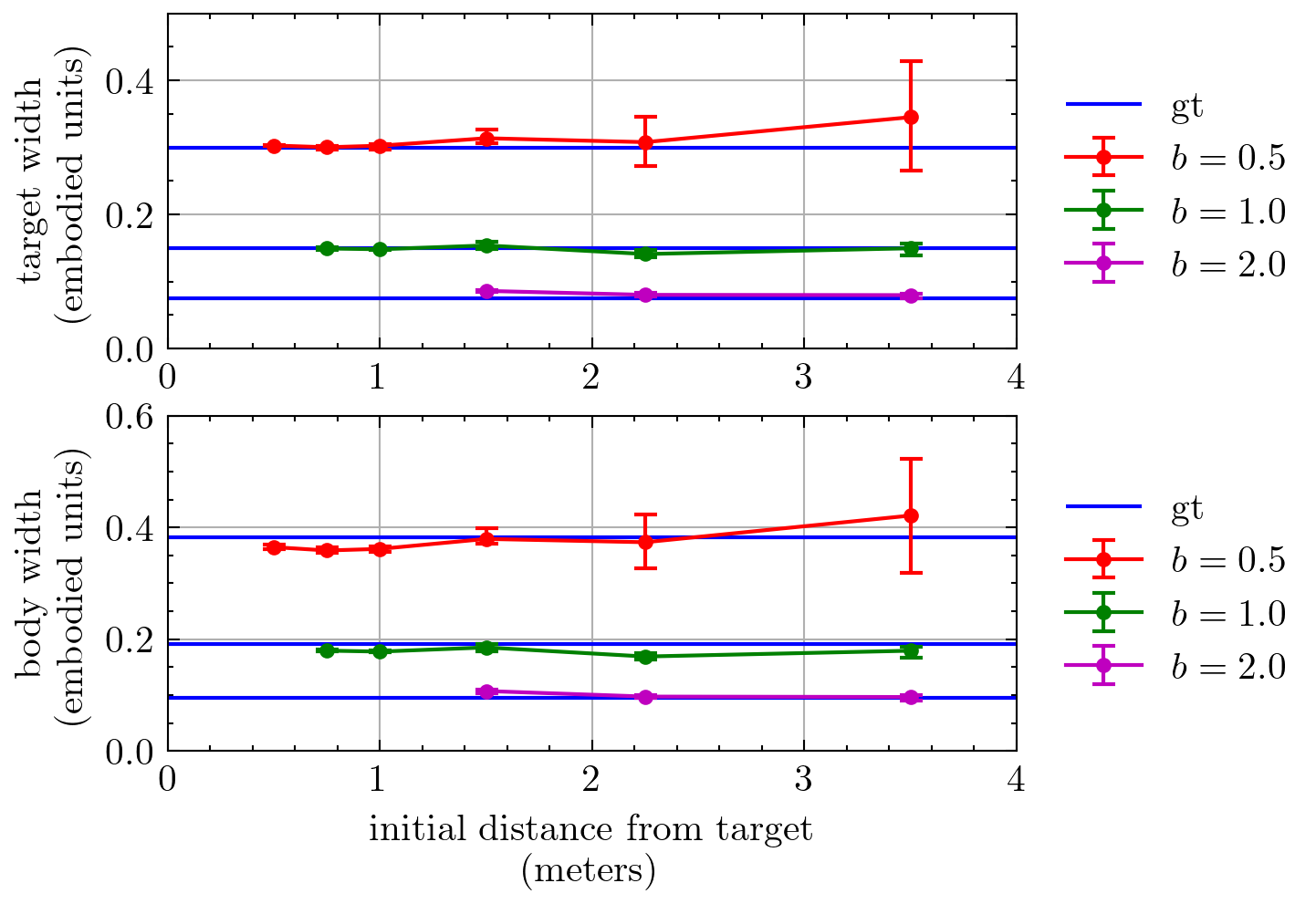}
  \caption{\blue{Measured width of touching target and width of the robot's body, in embodied units, as the initial distance from the target and the control input gain $b$ are varied. Error bars show mean, minimum, and maximum value over five trials.}}
  \label{fig:distancetarget}
\end{figure}

\blue{Additionally, two experiments were conducted to validate the accuracy of estimated embodied representations. In the first, the  uncalibrated touching procedure was performed with the robot starting from an initial distance of 50, 75, 100, 150, 225, and 350 cm away from a 15 centimeter wide touch target. At each distance, the input gain $b$ was set to 0.5, 1, and 2. When $b=$ 1 and $b=$ 2, the minimum initial distances were 75 cm and 150 cm, respectively, to avoid hitting the target during the open loop oscillation. In total, 70 trials were run. 350 cm was the maximum testable distance; beyond this, the procedure failed due to the limited width of the target in the field of view (\textless 6 pixels).
Figure \ref{fig:distancetarget} shows the resulting embodied target widths and embodied robot widths compared to ground truth. Table \ref{table:initialdistvar} quantifies the error in Figure \ref{fig:distancetarget} as a percentage of the ground truth target and body width.}

\begin{figure}
  \centering
  \includegraphics[width=0.5\linewidth]{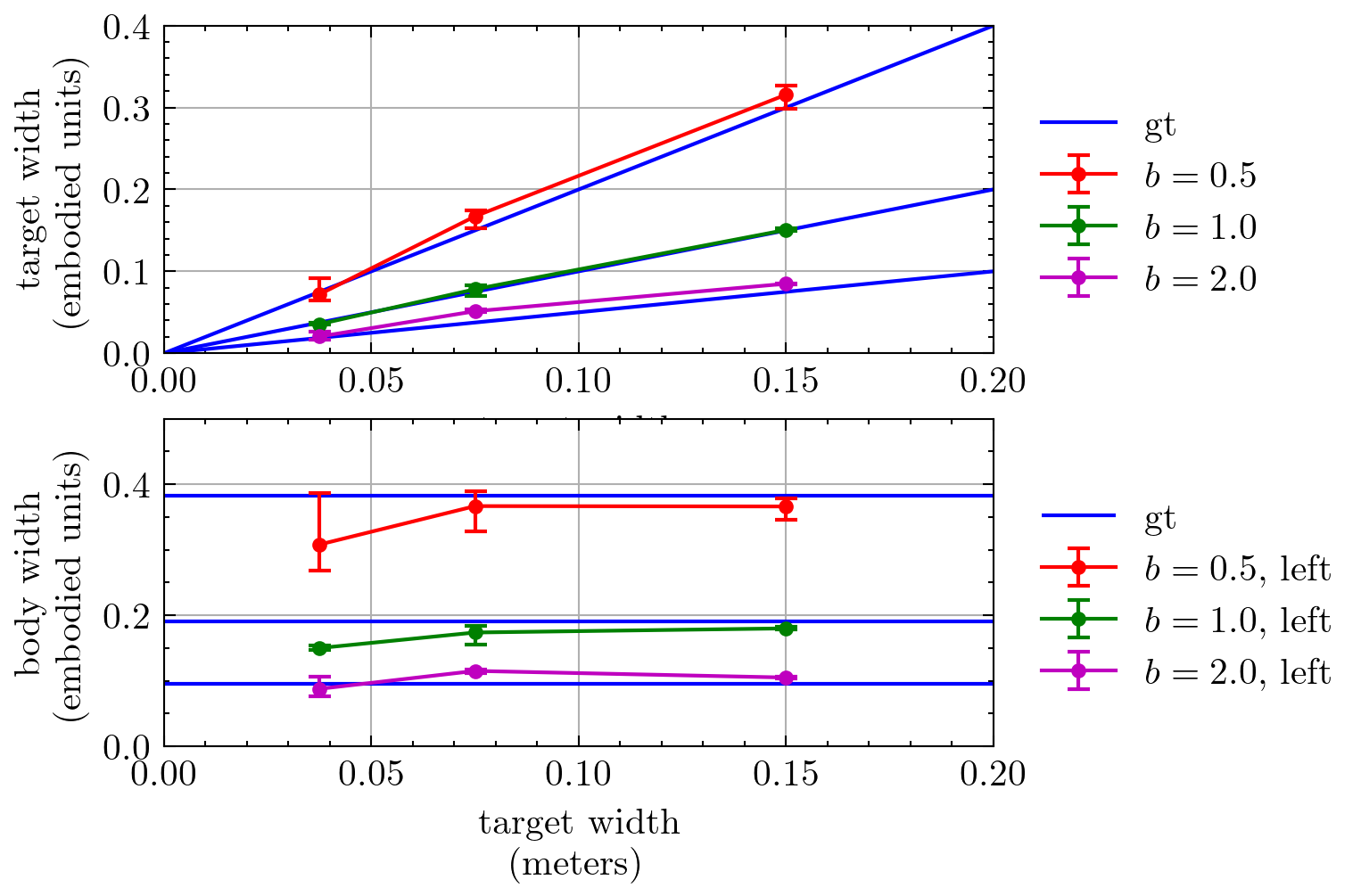}
  \caption{\blue{Measured width of touching target and width of the robot's body, in embodied units, as the width of the target and the control input gain $b$ are varied. Error bars show mean, minimum, and maximum value over five trials.}}
  \label{fig:widthtarget}
\end{figure}

\blue{In the second experiment, the uncalibrated touching procedure was performed with the robot initially 150 cm from targets of ground truth width 3.75, 7.5, and 15 cm and input gain $b$ equal to 0.5, 1, and 2. Five trials of each parameter combination were run, resulting in a total of 45 trials. Figure \ref{fig:widthtarget} shows the resulting embodied target widths and embodied robot widths compared to ground truth. The ground truth widths result from converting the width of the robot in meters minus the turning radius of the pan-tilt mount to the embodied distance. Table \ref{table:targetvar} quantifies the error in Figure \ref{fig:distancetarget} as a percentage of the ground truth target and body width.}

\blue{In Figures \ref{fig:distancetarget} and \ref{fig:widthtarget}, the blue ground truth lines result from converting the ground truth width of the target and width of the robot to meters. The plotted ground truth robot width was reduced by the turning radius of the pan-tilt mount that the camera is mounted on in embodied units.}

\blue{During all trials, the robot applied an open-loop 1/3 Hz acceleration signal, resulting in positional oscillation with 25 cm of amplitude when $b=$ 1. The sliding window estimator was configured to consider 3 seconds of data.}

\subsection*{Uncalibrated Clearing}

Once a robot can touch, it can quickly \blue{determine if it can} clear obstacles \blue{of initially unknown size prior to reaching them} as shown in Figure \ref{fig:clearing} and in Movie 1. \blue{In the following, the first four steps of and the estimation portion of step five are the same as in the touching application. Thus, we focus on the last portion of the fifth step, which is applying quantities estimated via Embodied Visuomotor Representation in the task.}

Let the robot approach the touching target along direction $v$ in the body frame; then, if the body extends further than the camera in the direction of $v$, the robot's body will touch the target prior to the camera reaching the target. Let the time of contact be called $t_c$. At \blue{the time of contact}, the \blue{vector} from the camera to the point on the body that is touching the target is \blue{known and corresponds with a point on the perimeter of the robot's body}. Repeating the touching process by approaching a target from many directions allows the robot to approximate its convex hull in the scale of the embodied units.

Clearing obstacles is now as simple as measuring the distance from the robot's camera to the object in the embodied unit and comparing that to the robot's size in the embodied unit. In particular, \blue{as illustrated by Figure \ref{fig:touching},} if a robot has measured its width \blue{in the embodied units as $w/b = w_l/b + w_r/b$, where $w_l/b$ and $w_r/b$ are distance from the camera to the left and right side of the robot respectively}, then it only needs to measure \blue{width of the opening, as illustrated in Figure \ref{fig:clearing},} and compare it to the width of the robot. Let the positions of the left and right sides of the opening be $X_l$ and $X_r$; then the robot measures their positions as $X_l/b$ and $X_r/b$. \blue{Thus,} the robot fits if

\begin{equation}
w < \|X_l - X_r| \iff \frac{w}{b} < \left\|\frac{X_l}{b} - \frac{X_r}{b}\right\|.
\end{equation}

\blue{If the robot fits, it can proceed through the opening using the same controller that was used to touch targets.}

\begin{figure}
  \centering
  \includegraphics[width=1.0\linewidth]{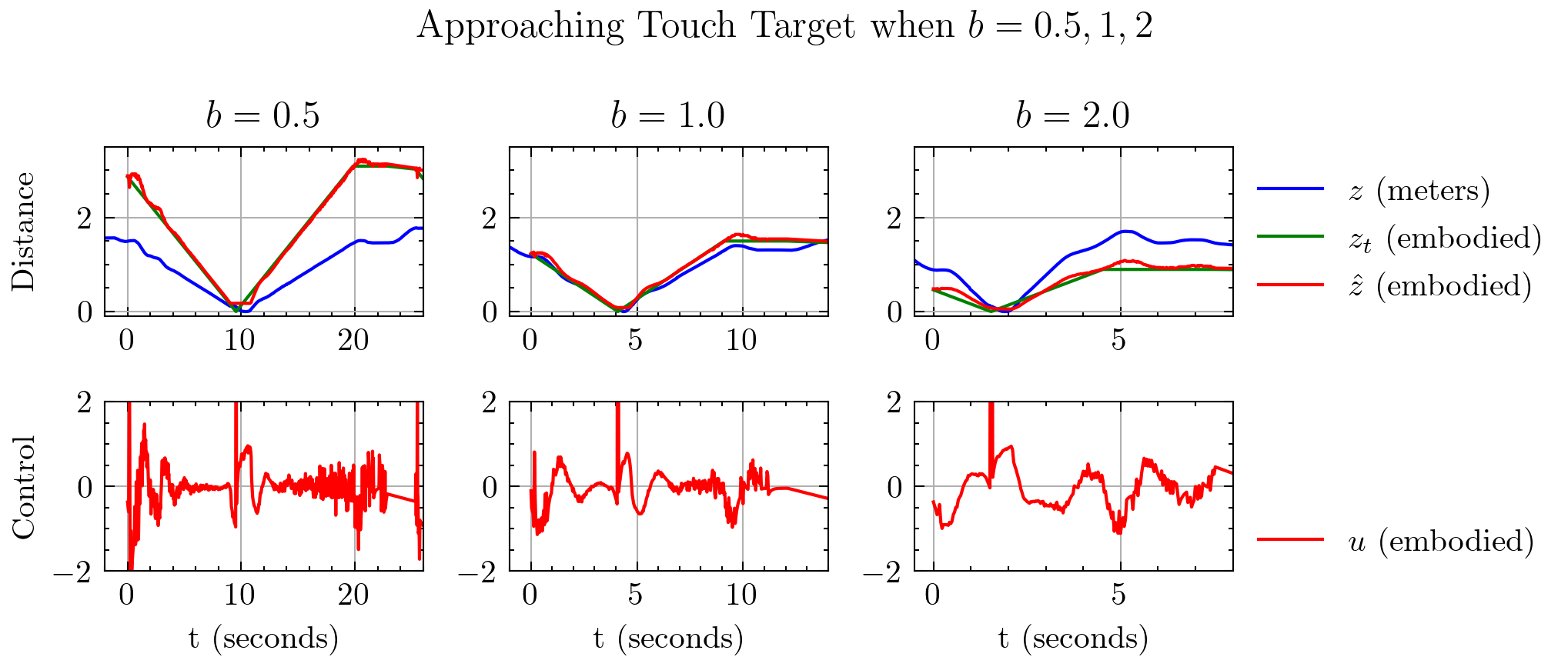}
  \caption{\blue{Experimental data and estimated signals from the target approach phase of uncalibrated touching for three different values of the control gain $b$. The methods take different amounts of time to reach the target because the approach speed was specified as a fixed value in the embodied unit while $b$ was varied. Note that the closed loop behavior remains stable and qualitatively similar despite the control input gain varying by a factor of 4. As shown by Equation \eqref{eqn:bcancelsdoubleint} this is expected.}}
  \label{fig:approachtarget}
\end{figure}

\blue{Figure \ref{fig:approachtarget} shows estimated distances from the opening as the robot approaches using different ground truth values for $b =$ 0.5, 1, and 2 as before. The approach corrects a translational perturbation in the X axis in addition to approaching the target at a constant speed along the Z axis. The approach velocity was fixed to the same embodied value for each trial; thus, when $b=$ 0.5 and $b=$ 2, the robot takes twice and half as long, respectively, to reach the target as when $b=$ 1.}

\begin{figure}
  \centering
  \includegraphics[width=0.5\linewidth]{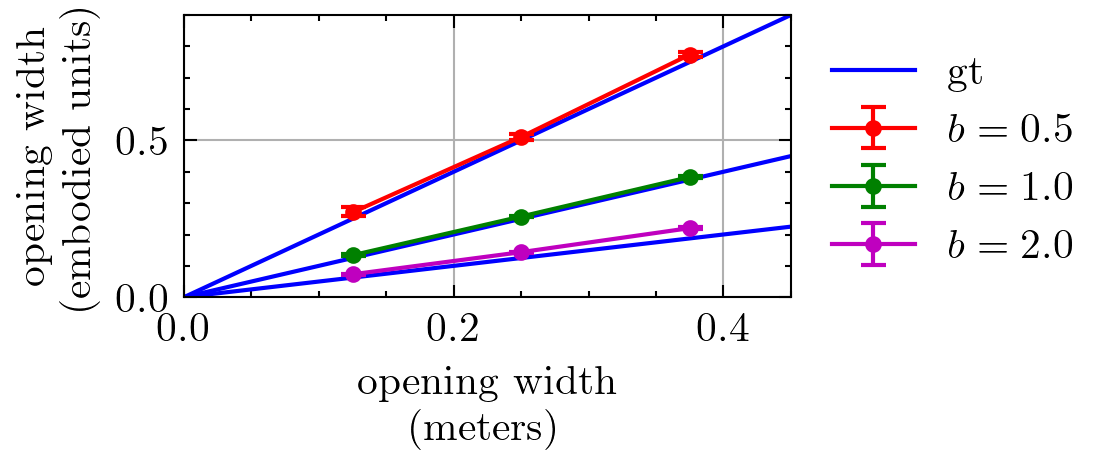}
  \caption{\blue{Measured width of opening width, in embodied units, as the width of the opening and the control input gain $b$ are varied. Error bars show mean, minimum, and maximum value over five trials.}}
  \label{fig:widthopening}
\end{figure}

\blue{Additionally, Figure \ref{fig:widthopening} shows the measured width of the opening as a function of the opening's true width. Openings of size 12.5, 25, and 37.5 cm were tested with the robot's initial distance from the opening fixed at 150 cm. Each configuration of opening size and control gain $b$ were tested 5 times for a total of 45 trials. It should be noted that measuring an opening's width is substantially similar to measuring a target's width, and so Figure \ref{fig:widthopening} can be interpreted as an extension of the upper plot in Figure \ref{fig:widthtarget}.  Table \ref{table:targetvar} quantifies the error in Figure \ref{fig:distancetarget} as a percentage of the width of the opening.}

\subsection*{Uncalibrated Jumping}

Suppose an uncalibrated robot with two legs, \blue{that apply force with an unknown first order response (or lag)}, and a monocular camera needs to jump a gap to a platform an unknown distance away but at the same height. \blue{In what follows, we carefully apply the five steps for using Embodied Visuomotor Representation. Figure \ref{fig:jumping} and Movie 2 of the Supplemental Material outline the uncalibrated jumping procedure visually.}

\blue{Starting with the first step, we assume a classical model of the system.} Let the legs applied force respond to inputs according to a first order low pass filter with a time constant $1/\alpha, \quad \alpha > 0$. That is,

\begin{equation}
\begin{split}
\ddot{X}\blue{_2} &= b x_{act} - g\blue{_b}\\
\dot{x}_{act} &= \alpha\left(u - x_{act}\right),
\end{split}
\end{equation}
where \blue{$\ddot{X}_2$ is acceleration in the vertical direction and} $g\blue{_b}$ is \blue{an unknown} gravitational bias. \blue{Then, as with the the touching example, all the quantities defined by the world, that is $b$, $g\blue{_b}$, and $\alpha$ do not have predetermined units or magnitudes. Thus, the second step is completed.}

\blue{To complete the third step, assume} the robot fixates on a \blue{horizontal line} that it needs to jump to and oscillates up and down. \blue{Then, the line's height in the image in normalized pixels can be used as $\Phi_W$. That is, $\Phi_W = X_2 / X_3$ where $X_2$ is the vertical position of the line and $X_3$ is the distance between the camera and the line. Then, the characteristic scale of vision $d$ is equal to $X_3$ and is the distance to be jumped.}

\blue{The fourth step of establishing a constraint between vision and control that involves the unknowns is accomplished by recalling Equation \eqref{eqn:generalizedphiconst}, specializing it to this example's dynamics, and adding the additional gravitational bias term. Finally, the fifth step can applied: estimating the force of gravity and the distance to the line in embodied units using} Equation \eqref{eqn:slidingadaptiveproblem}. \blue{The full formulation is given in the methods section and specifically Equation \eqref{eqn:jumpingproblem}. There, a family of truncated exponential decay functions are used to represent the first order actuator dynamics, which results in a linear least squares problem and a unique solution. Subsequently, Equation \eqref{eqn:dcgainscale} can be used to recover the distance to be jumped in the embodied units as $\hat{X}_3(t_0) / b$}. \blue{Figure \ref{fig:jumping} illustrates this procedure.}
 
Then, given the embodied jumping distance $d$, embodied gravity estimate, and a predetermined launch angle $\theta_l$, the robot can solve the projectile motion equations to determine a launch velocity\blue{, $v_l$,} and the landing time $t_f$. \blue{To find control inputs that achieve this launch velocity}, we \blue{pose a convex} optimal control problem, \eqref{eqn:jumpingcontrol}, \blue{that solves for a non-negative control input with minimum total variation that reaches the launch velocity. The formulation of both problems is given in the methods section.}

\begin{figure}
  \centering
  \includegraphics[width=1.0\linewidth]{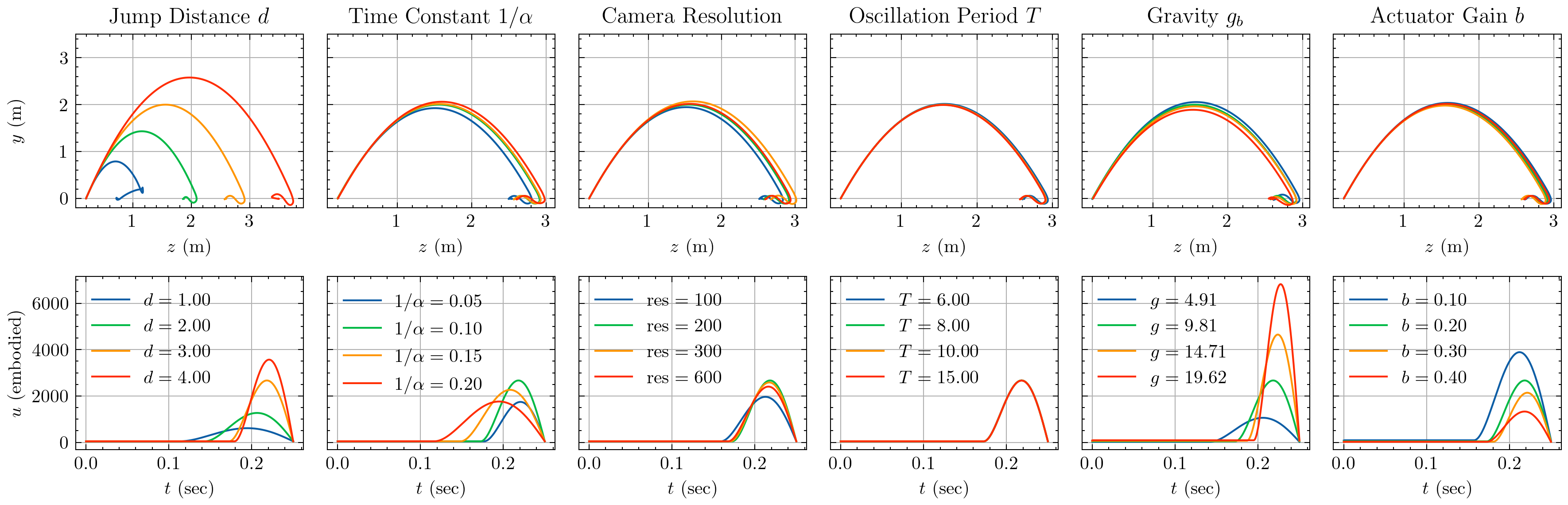}
  \caption{\blue{\textbf{Jump trajectories and jump control signal versus experimental parameters.} The procedure can successfully jump over a wide range of unknown gaps varying from 1 to 4 meters in width. A nominal gap of 3 meters can be jumped despite variations in the unknown actuator time constant, camera resolution, oscillation period, gravitational force, and actuator gain. The jump control signals resulting from an embodied optimal control problem \eqref{eqn:jumpingcontrol} can be seen to vary in magnitude and duration as expected given varied jump distance, actuator time constant, gravitational strength, and actuator gain.}}
  \label{fig:jumpingjumpdata}
\end{figure}

\blue{Figure \ref{fig:jumpingjumpdata} illustrates the results of this procedure in terms of the achieved jumping trajectory (in meters) and the applied control signal $u$ in embodied units. Figure \ref{fig:jumpingestimationdata} illustrates the accuracy of the estimate of the distance to be jumped, $d$, and the gravitational force, $g_b$, in the embodied unit after completing the measurement phase. Estimates resulting from processing simulated images with color thresholding and simulating worst case pixel quantization errors are shown. The error due to finite resolution can be seen to decrease by approximately $1/$resolution. A theoretical justification for this is given in the Supplemental Material. For each set of experiments, the non-varied parameter values were set to $d=3.0$, $1/\alpha = 0.1$, $\mathrm{res}=200$, $T=10$, $g_b=9.81$, and $b=0.2$.}

\blue{
It should be noted that performing the up and down oscillation requires a controller to compensate for the effects of gravity. For this experiment, the control gains were chosen manually, but they could easily be chosen automatically using traditional system identification and control design methods. Additionally, the mass of the robot's legs and feet have been neglected as the effect of their mass on the jumping trajectory cannot be estimated by oscillating the body alone. For the purpose of this paper, it suffices to assume the mass of the legs and feet is small, and so does not significantly affect the achieved jumping trajectory. However, the effect of the added mass of the legs and feet can be estimated by doing a vertical test jump prior to jumping the gap. Finally, because the approach is model based, it, in theory, works for all gap sizes, gravitational strengths, and actuator time constants. However, given fixed parameters, it will fail in specific scenarios where the model's assumptions do not hold. For example, if the strength of gravity is made very weak, the agent may fall off the jumping platform during the measurement phase unless the oscillation frequency and magnitude are also reduced.}
\subsection*{Comparisons to Bees and Gerbils}

Next, we contrast the properties of algorithms using Embodied Visual Scale with behavior observed in bees and gerbils.

Bees have been recently shown to understand the size of an opening in wingbeats during their approach to an opening. It is thought that the size of the wings is learned during early development through contact with the hive. Additionally, the bees oscillate \blue{increasingly vigorously} as openings are made smaller, seemingly to determine the size of the hole more accurately \cite{ravi2020bumblebees}. \blue{Similarly,} our method for uncalibrated clearing determines the size of the robot in embodied units through contact \blue{and oscillates to determine the size of the opening.} Additionally, \blue{an examination of the} least squares formulation \blue{admits} that increased oscillation over longer periods will increase the accuracy \blue{and robustness} of the estimated quantities \blue{because errors can be averaged out.}

Similarly, Mongolian Gerbils vibrate their body up and down before making a jump over a gap of unknown distance. It is thought that this helps their visuomotor system determine \blue{the} appropriate muscle stimulation to clear the gap \cite{ellard1984distance}. Like bees, the intensity of the oscillations increases when the gap is larger and thus is more difficult to jump over. Our approach also requires oscillation, can estimate the appropriate actuator stimulation to jump a gap, benefits from \blue{larger} oscillations \blue{(which induces a larger baseline for vision), and can increase its accuracy by considering longer timer intervals.}

\section*{Discussion}
\blue{Embodied Visuomotor Representation is a concrete methodology} that allows robots to emulate \blue{the process of estimating distance through action -- a capability well established in the natural sciences.} This \blue{approach} simplifies \blue{implementation by allowing} low-level control laws to self-tune and \blue{removing the need for detailed physical models or calibrated 3D sensors.} \blue{As a result, even uncalibrated robots can reliably } perform  tasks such as touching, clearing obstacles, or jumping, while maintaining guaranteed stability. 

\blue{Traditionally, body-length units have been proposed as a natural basis for embodied scale. However, such measurements are not directly accessible to an uncalibrated robot. Embodied Visuomotor Representation addresses this limitation by providing a mechanism through which a robot can internally estimate or re-estimate body-length units as needed. Instead of relying on external units like meters, it grounds the perception of distance using the robot’s own acceleration response to control inputs.}

We further compare traditional approaches based on sense-plan-act with our approach based on Embodied Visuomotor Representation in Table \ref{tab:embodimentmodes}. The table compares the main algorithmic components of traditional approaches that might be used to achieve touching, clearing, and jumping. This comparison reveals how the stability of traditional approaches becomes implicitly tied to \blue{prior} knowledge of the conversion factor between an embodied scale and an external scale \blue{due to the assumption of control inputs} on an external scale. \blue{Embodied Visuomotor Representation can escape this situation by using a different architecture from the traditional sense-plan-act cycle as illustrated by Figure \ref{fig:embodied}.}

The development of Embodied Visuomotor Representation has borrowed ideas broadly from many areas, in particular the Tau Theory, Glenberg's theory of embodiment as memory, the Internal Model \blue{Principle}, and Visual Self-Models. Below, these frameworks are compared and contrasted to Embodied Visuomotor Representation. We further compare our framework to leading approaches in robotics, including \blue{Direct Optical Flow Regulation, Visual Servoing,} Visual Inertial Odometry (VIO), Visual Self-Modeling, and Deep Visuomotor Control.


\blue{The {\it Tau Theory} originally proposed an embodied visuomotor relationship} as the basis for how time-to-contact could be used by humans for \blue{tasks such as driving a car} \cite{lee1976theory}.
This was developed into the {\it General Tau Theory} \cite{leegeneral2009}, which aims to explain numerous 
behaviors
using time-to-contact (tau).
\blue{Later theories generalized it to motion in all directions, called looming
\cite{raviv1992quantitative}. Tau Theory does not provide a specific methodology for achieving tau trajectories
and so roboticists have proposed methods ranging from direct regulation \cite{Sikorski2021, walters2021evreflex, herisse2012} to indirect estimation of position followed by regulation \cite{croon_2013, ho2017distance, TTCDist}, with most relying on pre-tuned of control laws to account for the unknown embodied scale.}
Embodied Visuomotor Representation can use time-to-contact as the basis for a scaleless visual transition matrix $\Phi_W$ as in \cite{TTCDist} without committing to any particular control law. Further, it is easy to use the framework to verify if a proposed control law is stable.

\blue{
A generalization of Tau Theory is {\it Direct Optical Flow Regulation}. Here, arbitrary optical flow fields are used as the setpoint to a control loop.
Such control schemes have been shown to result in striking similarities to insect flight behavior
\cite{ruffier2005, franceschini2007bio, serres2008vision, ruffier2015optic}.
Notably, \cite{de2022accommodating} showed that the framework allows orientation with respect to gravity to be estimated using oscillations of orientation.
Such schemes depend on pre-tuning of the control law for the expected heights and velocities.
\cite{de2015distance}
addresses this limitation by lowering the open-loop gain when
oscillations occur.
Embodied Visuomotor Representation can continuously estimate embodied distance without necessarily oscillating. Thus, it can be used to follow more general trajectories while simultaneously scaling the gains of a direct optical flow regulator to ensure stability.
}

\blue{
Another approach to visuomotor control is {\it Visual Servoing}
which considers objectives based on position (pose based) or image coordinates (image based)
\cite{Chaumette2006RAMa, Chaumette2006RAMb}.
When combined with Embodied Visuomotor Representation, pose based approaches no longer require a 3D model of known scale as demonstrated by uncalibrated touching and clearing.
This situation is similar to that considered by Image Based Visual Servoing, which operates without knowledge of scale by basing the control objective on the pixel positions and mitigating the effects of unknown depth using approximations, partitioning, or adaptivity \cite{Chaumette2006RAMa, Chaumette2006RAMb, Chen2005}.
Embodied Visuomotor Representation provides an estimate of depth, and so is closely related to adaptive visual servoing.
However, Embodied Visuomotor Representation is also a broader framework that enables a rich set
behaviors such as the examples of touching, clearing, and jumping.
}

\blue{Our implementation of these behaviors has strong parallels with} {\it Glenberg's theory of embodiment as memory} \cite{Glenberg_1997}. Glenberg writes, ``embodied representations do not need to be mapped onto the world to become meaningful because they arise from the world'' and our proposed Embodied Visuomotor Representation does just that.
A core tenet of Glenberg's theory is the ``merge'' operation that allows memories of different types of actions to be combined to do a new task. 
Our Embodied Visuomotor Representation admits a path \blue{towards realizing a general motion-based} ``merge'' operator because it explicitly represents the system's dynamics.

\blue{Realizing such advanced behavior requires the ability to predict. This is supported by t}he {\it Internal Model Principle}\blue{,}
that suggests an internal model of the process to be regulated must be contained within a controller if it is to succeed. This principle has analogs
in psychology, where it
is accepted that \blue{humans depend on} internal models
to control themselves
\cite{wolpert1995, Wolpert2019}. \blue{Further, recent evidence strongly suggests that even simple animals such as dragonflies have such predictive visuomotor models \cite{mischiati2015internal, huang2018}.}
\blue{In computer science,
{\it Animate Vision} was proposed
to explain how such visuomotor control
processes can be learned through interaction with the environment
\cite{BALLARD19923, BALLARD199157}.}
These theories are very general and do not explicitly consider the
\blue{challenges} introduced by the unit-less \blue{nature of 
vision.} Thus, Embodied Visuomotor Representation can be seen as a specialization \blue{that can realize versions of these theories in robots}.

\blue{An important component of predictive modeling is understanding the shape of the body. While the given examples of uncalibrated touching and clearing are the most basic way to do this, it is of interest to combine Embodied Visuomotor Representation and recent work on {\it Visual Self-Modeling} that consider neural representations of a robot's 3D form} \cite{chen_2022fullbody}.
Similarly, it is of interest to understand Embodied Visuomotor Representations application to {\it Deep Visuomotor Control} as
exemplified by \cite{levine2016end}.
Recent work has considered visual navigation, where a robot must learn to navigate through a scene from visual inputs and output concrete actions to be taken by low-level controllers \blue{provided by a variety of morphologies \cite{shah2021, shah2023, shah2023vint}.}

\blue{Next, we turn to {\it Visual Inertial Odometry} (VIO), and its relation to Embodied Visuomotor Representation. VIO is the conventional approach to estimating a robot's state from vision and calibrated inertial measurement units (IMUs).}
Typical approaches follow
a match, estimate, and predict architecture to \blue{estimate a metric trajectory and the 3D position of tracked points \cite{mourikis2007multi, bloesch2015robust, bloesch2017iterated, qin2018vins}.}
Architecturally, this is identical to sense-plan-act but in the estimation setting. 
In contrast, Embodied Visuomotor Representation \blue{generally} considers the characteristic scale of vision, which may not be the distance to features.
This encourages an object-centered visual framework where control tasks are considered with respect to individual objects.
\blue{Similarly, the role of IMUs is different within Embodied Visuomotor Representation.}
They are useful for
estimatating rotation 
over short periods.
and their acceleration can be used up to scale to better estimate mechanical parameters. They also allow a robot to \blue{continuously estimate} the conversion factor between its embodied scale and an external scale \blue{as might be needed to adhere to a safety standard or follow high level directions.}

The above comparisons to existing concepts resulted in several directions for future work.
However, there are many more, and so below, we outline the future work in theory and applications that we consider most important.

\blue{In terms of theory,} this paper has considered only equality-constrained Embodied Visuomotor Representation. That is, the unitless quantities from vision are considered to exactly equal some operator of control inputs. 
Ideas from psychology, and, in particular, the Tau Theory, suggest that more qualitative processes will suffice for many applications. Thus, it is of interest to develop inequality-constrained Embodied Visuomotor Representation in hopes of realizing qualitative representations and qualitative control laws that robots can use to accomplish tasks. \blue{Similarly, it is of interest to develop Embodied Visuomotor Representation in combination with direct methods such as Direct Optical Flow Regulation, and Image Based Visual Servoing. Here, an embodied representation of distance can be used to help initialize, learn, or select direct control laws. Finally, the role of rotation has recently been found to be useful for estimating attitude from direct measurements \cite{de2022accommodating}. Thus, it is of interest to carefully consider the role of rotation in embodied representation.}

Turning back to equality-constrained Embodied Visuomotor Representation, we have presented two types of estimation problems. The first, exemplified by \eqref{eqn:embodieddoubleintest}, estimates quantities in embodied units of action. The second formulation is given by \blue{\eqref{eqn:embodieddoubleintestunbiased} and} \eqref{eqn:slidingadaptiveproblem} \blue{which} estimate quantities in units of the visual processes characteristic scale. Subsequently, these quantities can be converted to an embodied unit of action. \blue{While the latter were shown to be unbiased with respect to zero mean noise in $\Phi_W$, it is still of interest to perform a more general and careful error analysis with the goal of constructing robust or optimal estimators for Embodied Visuomotor Representation.}

Next, we discuss \blue{the representation of} dynamics. The derivation of Embodied Visuomotor Representations above neglected damping and spring effects\blue{,} as might be caused by friction or tendon-like attachments, respectively. However, these effects can be incorporated, assuming these naturally occurring dynamics are stable, which they almost always will be. For example, we can incorporate into an Embodied Visuomotor Representation a system's dissipating energy due to spring and damping effects by considering velocity or position as the fundamental quantities controlled by the action instead of acceleration. In this paper, acceleration was used as the fundamental quantity controlled through action because \blue{all} animals and robots \blue{must use forces.}

Finally, it is interesting to consider more general dynamics than linear systems and how to control them via Embodied Visuomotor Representation. In principle, except for the use of unit-less visual feedback, this is not substantially different from existing work in robot dynamics, and we expect many techniques to port over directly. On the other hand, the linear dynamics considered in this paper should be sufficient for many applications. Thus, it is of interest to apply recent advancements in adaptive control, such as kernel methods that guarantee estimation of BIBO stable impulse responses \cite{bisiacco2020}.

\blue{Next we consider future work in applications. A straightforward application} that can be accomplished via Embodied Visuomotor Representation is avoiding and pursuing dynamic objects. \blue{If a position-based approach is used, the} basic procedure is identical to the touching, clearing, and jumping, except that an additional model of the dynamic object must be considered. \blue{Catching is also of interest. However, it is thought that humans use direct methods (that is, they do not primarily rely on 3D position) to catch objects. The evidence is particularly strong in sports where humans achieve extraordinary performance despite sensorimotor delays \cite{Seville1968, Regan1997}. Then, since Embodied Visuomotor Representation allows for learning predictive self models that could be used for delay compensation (as is thought to be essential in biological systems \cite{Wolpert2019, mischiati2015internal}), it is of particular interest to study the theoretical connections between embodied representation and direct methods in the task of robot catching}.

We also consider imitation and embodied affordance estimation to be the next two most important applications. Assuming the embodied mechanical and dynamics parameters have already been estimated, imitation within Embodied Visuomotor Representation becomes a problem of determining the embodied control signals that accomplish a visually observed behavior. The characteristic scale-based representation of vision will then naturally allow robots to imitate an action using the units of their own embodiment instead of an external scale.

Finally, while this paper has exclusively considered visual processes and their connection to motor signals, the basic principles should be extended to consider \blue{multiple modalities such as tactile sensing} and audio. We are particularly interested in \blue{tactile} sensing because \blue{grasping} is an action that \blue{results in both tactile and visual responses}.

\section*{Methods}
\subsection*{Hardware}
The robot platform used for the touching and clearing experiments is a DJI RoboMaster EP. The platform features mecanum wheels, which allow it to translate omnidirectionally. The provided robot arm and camera were removed and replaced with a pan-tilt servo mount and a Raspberry Pi Camera Module 3 with a \blue{120 degree} wide-angle lens \blue{running at $1536\times864$ resolution and 30 frames per second}. A Raspberry Pi \blue{5} interpreted images and calculated control commands for the robot. \blue{The associated code is provided in the Supplemental Material.}

The DJI RoboMaster does not support direct acceleration control. The lowest level supported is velocity control. To emulate the double integrator described in the touching and clearing experiments, the acceleration commands from the proposed control laws were transformed into velocity commands using a leaky integrator with a time constant of \blue{10} seconds.

\subsection*{Uncalibrated Touching}

The touching experiment uses color thresholding to detect the pixels of a planar target \blue{and calculate $\Phi_W$ based on the assumption of a flat target parallel with the image plane}. \blue{As mentioned in the results section, this visual representation was chosen as a pedagogical example and will not work for non-flat objects. \blue{In general,} the theory holds without modification for any other vision algorithm that provides position to scale as per \eqref{eqn:controlvisioncouple}.} The robot oscillates in open-loop by applying a \blue{1/3} Hz sinusoidal acceleration signal over a \blue{10} second interval. \blue{The amplitude of acceleration was set so that the positional oscillations had an amplitude of 25 cm when $b=1$.} Equation \eqref{eqn:embodieddoubleintestunbiased} is used to solve for the current position, velocity, and size of the touching target in embodied units using a \blue{3} second sliding window of measurements. \blu{To do so, the outermost integral of Equation \eqref{eqn:embodieddoubleintestunbiased} is approximated in discrete time, so that the problem can be considered as a standard linear least squares problem of the form $\min_x \|Ax-b\|^2$. This problem is transformed to the normal equations $A^T A x = A^Tb$ and $x$ is computed using JAXopt's \cite{blondel2022efficient} Cholesky decomposition based solver for linear systems. The code is available in the supplemental material.}

Subsequently, the robot drives toward the target at a constant velocity of \blue{0.3} embodied units per second \blue{using PID control and positional feedback from vision in embodied units. The approach continues for the time} required to traverse the initial distance between the robot and the target. \blue{Since the robot's body is closer to the target than the camera,} the robot makes contact with the target sometime during this interval. \blue{Then,} the minimum distance between the robot's camera and the target is the horizontal distance between the robot's camera and the edge of the robot that made contact with the target first.

\blue{The PID control loop was tuned manually and was accomplished with just a few trials because the closed-loop behavior does not depend on the units of distance or the geometry of the scene. Thus, reasonable initial values for the PID gains can be determined from a desired closed loop bandwidth.}

\subsection*{Uncalibrated Clearing}

The clearing experiments consider \blue{whether} the robot can fit through an opening (clear it) prior to reaching the opening. To do so, the robot determines the width of the body in the embodied unit by touching its left and right sides to a nearby object using the previously described procedure. It remains for the robot to measure the size of an opening in the embodied unit. Once again, the left and right sides of the opening are detected with color thresholding. The left and right edges of the right and left sides of the opening, respectively, are used as the characteristic scale of vision. \blue{As with the last example, this visual representation is pedagogical.} \blue{The same sinusoidal acceleration and sliding window estimator used during uncalibrated touching then} estimates the size of the opening in the embodied units. If the robot is smaller than the opening, it proceeds through with the same PID-based position control \blue{as used for uncalibrated touching.}

\subsection*{Uncalibrated Jumping}
The jumping robot experiment was performed in a MuJoCo simulation \cite{todorov2012mujoco}. Codes for this simulation are provided in the Supplemental Material.

The robot consists of a body on two legs and feet. The legs are modeled as cylinders or pistons, with first-order activation dynamics and a time constant of 0.1 seconds. The body mass was set to 10 kg, and the mass of the legs and feet amounted to 0.22 kg. Two angular position controllers hold the legs upright during the measurement phase of the experiment and tilt the body forward approximately 20 degrees during the jumping phase.

The measurement phase of the jumping experiment consists of three parts. In the first, the minimum/maximum position of the body is estimated by recording the initial position as the bottom position, increasing the upwards force until the body begins to lift, initializing a PID controller's integrator term with the force that overcomes gravity, and using PID control to maintain a constant upwards velocity until the top position is reached. Subsequently, the PID controller oscillates the body for \blue{$10$} seconds at $1$ Hz around the midpoint between the bottom and top positions with an amplitude equal to $1/4$ of the distance between the top and bottom positions. \blue{While this approach introduces an assumption of a pre-tuned PID controller to control the body position, this could be eliminated by applying traditional system identification techniques in the embodied units.}

During oscillation, color thresholding is used to determine the vertical normalized pixel coordinate corresponding to a red line placed in the middle of the target platform. The characteristic scale of vision then becomes the distance between the robot and the red line. Subsequently, the measurements of the control effort applied by the PID control and the vertical normalized pixel coordinate are used to solve for the leg actuator's impulse response, the distance between the robot and the red line, and the initial velocity.

\blue{The simulated camera has a resolution of $200\times200$ pixels and a 90 degree vertical field of view and runs at 60 fps. Estimation of the maximum possible errors due to pixel quantization, as illustrated in Figure \ref{fig:jumpingestimationdata}, was achieved by quantifying the ground truth pixel position of the red line to integer coordinates with random quantization boundaries according to the formula $p_sim = \mathrm{round}(p_{gt} + \Delta) - \Delta$ and varying the value of $\Delta$ between $[-0.5, 0.5]$ with 26 steps.}

To approximate the actuator impulse response, a linear combination of \blue{$50$} first-order impulse responses truncated to \blue{4} seconds with a \blue{DC gain of $1$} and time constants ranging from \blue{$0.008$} to \blue{$0.5$} seconds are considered. The resulting quadratic program to estimate the distance to the line, initial velocity, and actuator dynamics is then given by:

\begin{equation}\label{eqn:jumpingproblem}
\begin{split}
    &\min_{\frac{\dot{X}_2(0), g_b, c}{d}} \int_2^T \left(\Phi_W(t) - t\frac{\dot{X}_2(0)}{d} - \frac{1}{2}\frac{g_b}{d}t^2 - \frac{g}{d} * \int_0^t \int_0^\sigma  u(\sigma_2) d\sigma_2 d\sigma  \right)^2 dt \\
    & \frac{g(t)}{d} = \sum_i c_i e^{-t/\tau_i}\\
    \mathrm{s.t.} & \quad c_i >= 0 \quad \forall i
\end{split}
\end{equation}
where $\dot{X}_2(0)$ is the initial velocity on the robot camera's vertical axis, $g_b$ is the gravitational bias force, $c_i$ are the coefficients of the basis functions for $g$ and $\tau_i$ are the corresponding time-constants of the basis functions. The double integral \blue{and convolution are} computed using a $0.001$ second approximation, corresponding with the simulation's timestep. The outermost integral is approximated with a $1/60$'th of a second timestep, corresponding to the 60 fps simulated camera. The loss is evaluated over the interval $[\blue{4}, T]$ instead of $[0, T]$ to prevent initial conditions from contributing to the actuator dynamics through the \blue{4} second long impulse response $g$. Non-negativity of $c_i$ is necessary to prevent overfitting to numerical integration errors. \blu{Similarly to the uncalibrated clearing and touching implementation, the problem is solved by approximating the outermost integral in discrete time. This results in a constrained quadratic program, which is solved using CVXOPT's quadratic program solver \cite{cvxoptbook}. The code is available in the supplemental material.} \blue{By solving this problem, the dynamics of the system are estimated automatically.}

To convert from units of the characteristic scale to embodied units of action, we divide each estimated quantity by $\int_0^{\infty}g(t)dt / d$. In what follows, all quantities can be assumed to be in this embodied unit.

Subsequently, the standard equations for projectile motion are combined with the estimated embodied distance and gravitational force to solve for an embodied launch velocity \blue{$v_l$} that will jump the gap. The equations to be solved are

\begin{equation}
\begin{split}
0 &= t_f \sin(\theta_l)\blue{v}_l - \frac{1}{2}t_f^2 g\\
d &= t_f \cos(\theta_l)\blue{v}_l,
\end{split}
\end{equation}

\blue{where $t_f$ is the time of landing.}

\blue{Finally,} A terminally constrained optimal control problem is solved to find the control input with a minimum \blue{total variation} that accelerates the body from rest to the launch velocity \blue{in 250 milliseconds over the distance oscillated over during the measurement phase.} The resulting quadratic program is:

\begin{equation}\label{eqn:jumpingcontrol}
\begin{split}
    &\min_u \int_0^T \left(\frac{d}{dt} u(t)\right)^2 dt \\
    \mathrm{s.t.} &\int_0^T (g * u)(t) dt = v_l \\
    & \int_0^T \int_0^\sigma (g * u)(\sigma_2) d\sigma_2 d\sigma = d_m \\
    &u(0) = 0, \quad u(T) = 0, \quad u(t) >= 0, \quad \forall t \in [0, T]
\end{split}
\end{equation}
where $v_l$ is the necessary launch velocity in embodied units, and $d_m$ is the distance from the start position at which the launch velocity should be achieved. $d_m$ is necessary to ensure the launch velocity is reached prior to reaching the maximum position of the body with respect to the legs.

Non-negativity of $u$ is required to prevent lifting the lightweight legs from the jumping platform prior to reaching the launch velocity. We formulate and solve the problem using a discrete time approximation of $0.001$ seconds. Because of exponential decay, $(g * u)(t) > 0$ when $t > T$\blue{,} and so the leg actuators are deactivated for $t > T$ to prevent the \blue{body's} velocity from exceeding $v_l$. The resulting embodied jump control signal $u$ is offset by the estimated gravitational bias and executed in an open loop.

\blue{Thus, by combining \eqref{eqn:jumpingproblem} and \eqref{eqn:jumpingcontrol}, the estimation of system dynamics and the synthesis of the jumping control input are fully automated, except for the use of a PID controller during the oscillation phase. However, this step could also be automated using existing system identification or adaptive control techniques. It is important to note that this example is pedagogical and does not account for or attempt to mitigate disturbances. Nonetheless, this does not affect the generality of the result, as Embodied Visuomotor Representation solely determines the units of distance and remains compatible with any jumping algorithm.}

\section*{Data Availability}
All data needed to interpret the conclusions of the paper are presented in the paper or Supplementary Materials. Raw datasets generated and analyzed during the study are available from the corresponding author upon reasonable request.

\section*{Code Availability}
Code for running the touching, clearing, and jumping experiments is provided in the Supplementary Materials and at \url{prg.cs.umd.edu/EVR}. Code for analyzing raw datasets collected during the study are available from the corresponding author upon reasonable request.

\section*{Acknowledgments}
The support of NSF award OISE 2020624 is gratefully acknowledged. The authors thank Rohit Kommuru for prototyping early versions of the jumping experiment.

\section*{Author Contributions}
L.B. conceived and developed the mathematics for Embodied Visuomotor Representation, designed the algorithms solving the uncalibrated touching, clearing, and jumping problems, built the robot,  wrote software, performed experiments, and wrote the manuscript. C.F. connected the ideas to existing literature and helped write the manuscript. Y.A. posed the questions answered by Embodied Visuomotor Representation, connected it to existing literature, and helped write the manuscript. All authors read and approved the final manuscript.

\section*{Competing Interests}
The authors declare no competing interests.



\bibliography{scifile}

\begin{thebibliography}{10}
\expandafter\ifx\csname url\endcsname\relax
  \def\url#1{\texttt{#1}}\fi
\expandafter\ifx\csname urlprefix\endcsname\relax\def\urlprefix{URL }\fi
\providecommand{\bibinfo}[2]{#2}
\providecommand{\eprint}[2][]{\url{#2}}

\bibitem{marr1982}
\bibinfo{author}{Marr, D.}
\newblock \emph{\bibinfo{title}{Vision: A computational investigation into the human representation and processing of visual information}} (\bibinfo{publisher}{MIT press}, \bibinfo{year}{2010}).

\bibitem{aloimonos1988active}
\bibinfo{author}{Aloimonos, J.}, \bibinfo{author}{Weiss, I.} \& \bibinfo{author}{Bandyopadhyay, A.}
\newblock \bibinfo{title}{Active vision}.
\newblock \emph{\bibinfo{journal}{International Journal of Computer Vision}} \textbf{\bibinfo{volume}{1}}, \bibinfo{pages}{333--356} (\bibinfo{year}{1988}).

\bibitem{bajcsy2018revisiting}
\bibinfo{author}{Bajcsy, R.}, \bibinfo{author}{Aloimonos, Y.} \& \bibinfo{author}{Tsotsos, J.~K.}
\newblock \bibinfo{title}{{Revisiting active perception}}.
\newblock \emph{\bibinfo{journal}{Autonomous Robots}} \textbf{\bibinfo{volume}{42}}, \bibinfo{pages}{177--196} (\bibinfo{year}{2018}).

\bibitem{BALLARD199157}
\bibinfo{author}{Ballard, D.~H.}
\newblock \bibinfo{title}{Animate vision}.
\newblock \emph{\bibinfo{journal}{Artificial Intelligence}} \textbf{\bibinfo{volume}{48}}, \bibinfo{pages}{57--86} (\bibinfo{year}{1991}).

\bibitem{nasa1999mars}
\bibinfo{author}{Stephenson, A.~G.} \emph{et~al.}
\newblock \emph{\bibinfo{title}{Mars Climate Orbiter Mishap Investigation Board Phase {I} Report}} (\bibinfo{publisher}{National Aeronautics and Space Administration}, \bibinfo{year}{1999}).

\bibitem{feldman1985four}
\bibinfo{author}{Feldman, J.~A.}
\newblock \bibinfo{title}{Four frames suffice: A provisional model of vision and space}.
\newblock \emph{\bibinfo{journal}{Behavioral and Brain Sciences}} \textbf{\bibinfo{volume}{8}}, \bibinfo{pages}{265--289} (\bibinfo{year}{1985}).

\bibitem{koenderink1992surface}
\bibinfo{author}{Koenderink, J.~J.}, \bibinfo{author}{Van~Doorn, A.~J.} \& \bibinfo{author}{Kappers, A.~M.}
\newblock \bibinfo{title}{Surface perception in pictures}.
\newblock \emph{\bibinfo{journal}{Perception \& Psychophysics}} \textbf{\bibinfo{volume}{52}}, \bibinfo{pages}{487--496} (\bibinfo{year}{1992}).

\bibitem{todd2004visual}
\bibinfo{author}{Todd, J.~T.}
\newblock \bibinfo{title}{The visual perception of {3D} shape}.
\newblock \emph{\bibinfo{journal}{Trends in cognitive sciences}} \textbf{\bibinfo{volume}{8}}, \bibinfo{pages}{115--121} (\bibinfo{year}{2004}).

\bibitem{cheong1998effects}
\bibinfo{author}{Cheong, L.}, \bibinfo{author}{Ferm{\"u}ller, C.} \& \bibinfo{author}{Aloimonos, Y.}
\newblock \bibinfo{title}{Effects of errors in the viewing geometry on shape estimation}.
\newblock \emph{\bibinfo{journal}{Computer Vision and Image Understanding}} \textbf{\bibinfo{volume}{71}}, \bibinfo{pages}{356--372} (\bibinfo{year}{1998}).

\bibitem{ji2006noise}
\bibinfo{author}{Ji, H.} \& \bibinfo{author}{Ferm{\"u}ller, C.}
\newblock \bibinfo{title}{Noise causes slant underestimation in stereo and motion}.
\newblock \emph{\bibinfo{journal}{Vision Research}} \textbf{\bibinfo{volume}{46}}, \bibinfo{pages}{3105--3120} (\bibinfo{year}{2006}).

\bibitem{ellard1984distance}
\bibinfo{author}{Ellard, C.~G.}, \bibinfo{author}{Goodale, M.~A.} \& \bibinfo{author}{Timney, B.}
\newblock \bibinfo{title}{Distance estimation in the mongolian gerbil: The role of dynamic depth cues}.
\newblock \emph{\bibinfo{journal}{Behavioural brain research}} \textbf{\bibinfo{volume}{14}}, \bibinfo{pages}{29--39} (\bibinfo{year}{1984}).

\bibitem{mischiati2015internal}
\bibinfo{author}{Mischiati, M.} \emph{et~al.}
\newblock \bibinfo{title}{Internal models direct dragonfly interception steering}.
\newblock \emph{\bibinfo{journal}{Nature}} \textbf{\bibinfo{volume}{517}}, \bibinfo{pages}{333--338} (\bibinfo{year}{2015}).

\bibitem{ravi2020bumblebees}
\bibinfo{author}{Ravi, S.} \emph{et~al.}
\newblock \bibinfo{title}{Bumblebees perceive the spatial layout of their environment in relation to their body size and form to minimize inflight collisions}.
\newblock \emph{\bibinfo{journal}{Proceedings of the National Academy of Sciences}} \textbf{\bibinfo{volume}{117}}, \bibinfo{pages}{31494--31499} (\bibinfo{year}{2020}).

\bibitem{lee1976theory}
\bibinfo{author}{Lee, D.~N.}
\newblock \bibinfo{title}{A theory of visual control of braking based on information about time-to-collision}.
\newblock \emph{\bibinfo{journal}{Perception}} \textbf{\bibinfo{volume}{5}}, \bibinfo{pages}{437--459} (\bibinfo{year}{1976}).

\bibitem{ruffier2005}
\bibinfo{author}{Ruffier, F.} \& \bibinfo{author}{Franceschini, N.}
\newblock \bibinfo{title}{Optic flow regulation: the key to aircraft automatic guidance}.
\newblock \emph{\bibinfo{journal}{Robotics and Autonomous Systems}} \textbf{\bibinfo{volume}{50}}, \bibinfo{pages}{177--194} (\bibinfo{year}{2005}).

\bibitem{Chaumette2006RAMa}
\bibinfo{author}{Chaumette, F.} \& \bibinfo{author}{Hutchinson, S.}
\newblock \bibinfo{title}{Visual servo control. {I}. {B}asic approaches}.
\newblock \emph{\bibinfo{journal}{IEEE Robotics \& Automation Magazine}} \textbf{\bibinfo{volume}{13}}, \bibinfo{pages}{82--90} (\bibinfo{year}{2006}).

\bibitem{Chaumette2006RAMb}
\bibinfo{author}{Chaumette, F.} \& \bibinfo{author}{Hutchinson, S.}
\newblock \bibinfo{title}{Visual servo control. {II}. {A}dvanced approaches}.
\newblock \emph{\bibinfo{journal}{IEEE Robotics \& Automation Magazine}} \textbf{\bibinfo{volume}{14}}, \bibinfo{pages}{109--118} (\bibinfo{year}{2007}).

\bibitem{herisse2012}
\bibinfo{author}{Herissé, B.}, \bibinfo{author}{Hamel, T.}, \bibinfo{author}{Mahony, R.} \& \bibinfo{author}{Russotto, F.-X.}
\newblock \bibinfo{title}{Landing a {VTOL} unmanned aerial vehicle on a moving platform using optical flow}.
\newblock \emph{\bibinfo{journal}{IEEE Transactions on Robotics}} \textbf{\bibinfo{volume}{28}}, \bibinfo{pages}{77--89} (\bibinfo{year}{2012}).

\bibitem{de2022accommodating}
\bibinfo{author}{De~Croon, G.~C.} \emph{et~al.}
\newblock \bibinfo{title}{Accommodating unobservability to control flight attitude with optic flow}.
\newblock \emph{\bibinfo{journal}{Nature}} \textbf{\bibinfo{volume}{610}}, \bibinfo{pages}{485--490} (\bibinfo{year}{2022}).

\bibitem{Chaumette1999}
\bibinfo{author}{Malis, E.}, \bibinfo{author}{Chaumette, F.} \& \bibinfo{author}{Boudet, S.}
\newblock \bibinfo{title}{2 1/2 {D} visual servoing}.
\newblock \emph{\bibinfo{journal}{IEEE Transactions on Robotics and Automation}} \textbf{\bibinfo{volume}{15}}, \bibinfo{pages}{238--250} (\bibinfo{year}{1999}).

\bibitem{Corke2001}
\bibinfo{author}{Corke, P.} \& \bibinfo{author}{Hutchinson, S.}
\newblock \bibinfo{title}{A new partitioned approach to image-based visual servo control}.
\newblock \emph{\bibinfo{journal}{IEEE Transactions on Robotics and Automation}} \textbf{\bibinfo{volume}{17}}, \bibinfo{pages}{507--515} (\bibinfo{year}{2001}).

\bibitem{Glenberg_1997}
\bibinfo{author}{Glenberg, A.~M.}
\newblock \bibinfo{title}{What memory is for}.
\newblock \emph{\bibinfo{journal}{Behavioral and Brain Sciences}} \textbf{\bibinfo{volume}{20}}, \bibinfo{pages}{1–19} (\bibinfo{year}{1997}).

\bibitem{brock2024intelligencecomputation}
\bibinfo{author}{Brock, O.}
\newblock \bibinfo{title}{Intelligence as computation}.
\newblock \emph{\bibinfo{journal}{IOP Conference Series: Materials Science and Engineering}} \textbf{\bibinfo{volume}{1321}}, \bibinfo{pages}{012001} (\bibinfo{year}{2024}).

\bibitem{Wolpert2019}
\bibinfo{author}{McNamee, D.} \& \bibinfo{author}{Wolpert, D.~M.}
\newblock \bibinfo{title}{Internal models in biological control}.
\newblock \emph{\bibinfo{journal}{Annual Review of Control, Robotics, and Autonomous Systems}} \textbf{\bibinfo{volume}{2}}, \bibinfo{pages}{339--364} (\bibinfo{year}{2019}).

\bibitem{huang2018}
\bibinfo{author}{Huang, J.} \emph{et~al.}
\newblock \bibinfo{title}{Internal models in control, biology and neuroscience}.
\newblock In \emph{\bibinfo{booktitle}{2018 IEEE Conference on Decision and Control (CDC)}}, \bibinfo{pages}{5370--5390} (\bibinfo{year}{2018}).

\bibitem{baker2004lucas}
\bibinfo{author}{Baker, S.} \& \bibinfo{author}{Matthews, I.}
\newblock \bibinfo{title}{{L}ucas-{K}anade 20 years on: A unifying framework}.
\newblock \emph{\bibinfo{journal}{International Journal of Computer Vision}} \textbf{\bibinfo{volume}{56}}, \bibinfo{pages}{221--255} (\bibinfo{year}{2004}).

\bibitem{astrom2008adaptive}
\bibinfo{author}{{\AA}str{\"o}m, K.} \& \bibinfo{author}{Wittenmark, B.}
\newblock \emph{\bibinfo{title}{Adaptive Control}} (\bibinfo{publisher}{Dover Publications}, \bibinfo{year}{2008}).

\bibitem{TTCDist}
\bibinfo{author}{Burner, L.}, \bibinfo{author}{Sanket, N.~J.}, \bibinfo{author}{Fermüller, C.} \& \bibinfo{author}{Aloimonos, Y.}
\newblock \bibinfo{title}{{TTCDist}: Fast distance estimation from an active monocular camera using time-to-contact}.
\newblock In \emph{\bibinfo{booktitle}{2023 IEEE International Conference on Robotics and Automation (ICRA)}}, \bibinfo{pages}{4909--4915} (\bibinfo{year}{2023}).

\bibitem{Hespanha2018}
\bibinfo{author}{Hespanha, J.~P.}
\newblock \emph{\bibinfo{title}{Linear Systems Theory}} (\bibinfo{publisher}{Princeton University Press}, \bibinfo{year}{2018}), \bibinfo{edition}{2nd} edn.

\bibitem{leegeneral2009}
\bibinfo{author}{Lee, D.~N.}, \bibinfo{author}{Bootsma, R.~J.}, \bibinfo{author}{Land, M.}, \bibinfo{author}{Regan, D.} \& \bibinfo{author}{Gray, R.}
\newblock \bibinfo{title}{Lee's 1976 paper}.
\newblock \emph{\bibinfo{journal}{Perception}} \textbf{\bibinfo{volume}{38}}, \bibinfo{pages}{837--858} (\bibinfo{year}{2009}).

\bibitem{raviv1992quantitative}
\bibinfo{author}{Raviv, D.}
\newblock \emph{\bibinfo{title}{A quantitative approach to looming}} (\bibinfo{publisher}{US Department of Commerce, National Institute of Standards and Technology}, \bibinfo{year}{1992}).

\bibitem{Sikorski2021}
\bibinfo{author}{Sikorski, O.}, \bibinfo{author}{Izzo, D.} \& \bibinfo{author}{Meoni, G.}
\newblock \bibinfo{title}{Event-based spacecraft landing using time-to-contact}.
\newblock In \emph{\bibinfo{booktitle}{Proceedings of the IEEE/CVF Conference on Computer Vision and Pattern Recognition (CVPR) Workshops}}, \bibinfo{pages}{1941--1950} (\bibinfo{year}{2021}).

\bibitem{walters2021evreflex}
\bibinfo{author}{Walters, C.} \& \bibinfo{author}{Hadfield, S.}
\newblock \bibinfo{title}{{EVR}eflex: Dense time-to-impact prediction for event-based obstacle avoidance}.
\newblock In \emph{\bibinfo{booktitle}{2021 IEEE/RSJ International Conference on Intelligent Robots and Systems (IROS)}}, \bibinfo{pages}{1304--1309} (\bibinfo{year}{2021}).

\bibitem{croon_2013}
\bibinfo{author}{Izzo, D.} \& \bibinfo{author}{Croon, G.}
\newblock \bibinfo{title}{Nonlinear model predictive control applied to vision-based spacecraft landing}.
\newblock In \emph{\bibinfo{booktitle}{Proceedings of the EuroGNC 2013, 2nd CEAS Specialist Conference on Guidance, Navigation \& Control}}, \bibinfo{pages}{91--107} (\bibinfo{year}{2013}).

\bibitem{ho2017distance}
\bibinfo{author}{Ho, H.~W.}, \bibinfo{author}{de~Croon, G.~C.} \& \bibinfo{author}{Chu, Q.}
\newblock \bibinfo{title}{Distance and velocity estimation using optical flow from a monocular camera}.
\newblock \emph{\bibinfo{journal}{International Journal of Micro Air Vehicles}} \textbf{\bibinfo{volume}{9}}, \bibinfo{pages}{198--208} (\bibinfo{year}{2017}).

\bibitem{franceschini2007bio}
\bibinfo{author}{Franceschini, N.}, \bibinfo{author}{Ruffier, F.} \& \bibinfo{author}{Serres, J.}
\newblock \bibinfo{title}{A bio-inspired flying robot sheds light on insect piloting abilities}.
\newblock \emph{\bibinfo{journal}{Current Biology}} \textbf{\bibinfo{volume}{17}}, \bibinfo{pages}{329--335} (\bibinfo{year}{2007}).

\bibitem{serres2008vision}
\bibinfo{author}{Serres, J.}, \bibinfo{author}{Dray, D.}, \bibinfo{author}{Ruffier, F.} \& \bibinfo{author}{Franceschini, N.}
\newblock \bibinfo{title}{A vision-based autopilot for a miniature air vehicle: joint speed control and lateral obstacle avoidance}.
\newblock \emph{\bibinfo{journal}{Autonomous robots}} \textbf{\bibinfo{volume}{25}}, \bibinfo{pages}{103--122} (\bibinfo{year}{2008}).

\bibitem{ruffier2015optic}
\bibinfo{author}{Ruffier, F.} \& \bibinfo{author}{Franceschini, N.}
\newblock \bibinfo{title}{Optic flow regulation in unsteady environments: A tethered mav achieves terrain following and targeted landing over a moving platform}.
\newblock \emph{\bibinfo{journal}{Journal of Intelligent \& Robotic Systems}} \textbf{\bibinfo{volume}{79}}, \bibinfo{pages}{275--293} (\bibinfo{year}{2015}).

\bibitem{de2015distance}
\bibinfo{author}{de~Croon, G. C. H.~E.}
\newblock \bibinfo{title}{Monocular distance estimation with optical flow maneuvers and efference copies: a stability-based strategy}.
\newblock \emph{\bibinfo{journal}{Bioinspiration {\&} Biomimetics}} \textbf{\bibinfo{volume}{11}}, \bibinfo{pages}{016004} (\bibinfo{year}{2016}).

\bibitem{Chen2005}
\bibinfo{author}{Chen, J.}, \bibinfo{author}{Dawson, D.}, \bibinfo{author}{Dixon, W.} \& \bibinfo{author}{Behal, A.}
\newblock \bibinfo{title}{Adaptive homography-based visual servo tracking for a fixed camera configuration with a camera-in-hand extension}.
\newblock \emph{\bibinfo{journal}{IEEE Transactions on Control Systems Technology}} \textbf{\bibinfo{volume}{13}}, \bibinfo{pages}{814--825} (\bibinfo{year}{2005}).

\bibitem{wolpert1995}
\bibinfo{author}{Wolpert, D.~M.}, \bibinfo{author}{Ghahramani, Z.} \& \bibinfo{author}{Jordan, M.~I.}
\newblock \bibinfo{title}{An internal model for sensorimotor integration}.
\newblock \emph{\bibinfo{journal}{Science}} \textbf{\bibinfo{volume}{269}}, \bibinfo{pages}{1880--1882} (\bibinfo{year}{1995}).

\bibitem{BALLARD19923}
\bibinfo{author}{Ballard, D.~H.} \& \bibinfo{author}{Brown, C.~M.}
\newblock \bibinfo{title}{Principles of animate vision}.
\newblock \emph{\bibinfo{journal}{CVGIP: Image Understanding}} \textbf{\bibinfo{volume}{56}}, \bibinfo{pages}{3--21} (\bibinfo{year}{1992}).

\bibitem{chen_2022fullbody}
\bibinfo{author}{Chen, B.}, \bibinfo{author}{Kwiatkowski, R.}, \bibinfo{author}{Vondrick, C.} \& \bibinfo{author}{Lipson, H.}
\newblock \bibinfo{title}{Fully body visual self-modeling of robot morphologies}.
\newblock \emph{\bibinfo{journal}{Science Robotics}} \textbf{\bibinfo{volume}{7}}, \bibinfo{pages}{eabn1944} (\bibinfo{year}{2022}).

\bibitem{levine2016end}
\bibinfo{author}{Levine, S.}, \bibinfo{author}{Finn, C.}, \bibinfo{author}{Darrell, T.} \& \bibinfo{author}{Abbeel, P.}
\newblock \bibinfo{title}{End-to-end training of deep visuomotor policies}.
\newblock \emph{\bibinfo{journal}{The Journal of Machine Learning Research}} \textbf{\bibinfo{volume}{17}}, \bibinfo{pages}{1334--1373} (\bibinfo{year}{2016}).

\bibitem{shah2021}
\bibinfo{author}{Shah, D.}, \bibinfo{author}{Eysenbach, B.}, \bibinfo{author}{Kahn, G.}, \bibinfo{author}{Rhinehart, N.} \& \bibinfo{author}{Levine, S.}
\newblock \bibinfo{title}{{ViNG}: Learning open-world navigation with visual goals}.
\newblock In \emph{\bibinfo{booktitle}{2021 IEEE International Conference on Robotics and Automation (ICRA)}}, \bibinfo{pages}{13215--13222} (\bibinfo{year}{2021}).

\bibitem{shah2023}
\bibinfo{author}{Shah, D.}, \bibinfo{author}{Sridhar, A.}, \bibinfo{author}{Bhorkar, A.}, \bibinfo{author}{Hirose, N.} \& \bibinfo{author}{Levine, S.}
\newblock \bibinfo{title}{{GNM}: A general navigation model to drive any robot}.
\newblock In \emph{\bibinfo{booktitle}{2023 IEEE International Conference on Robotics and Automation (ICRA)}}, \bibinfo{pages}{7226--7233} (\bibinfo{year}{2023}).

\bibitem{shah2023vint}
\bibinfo{author}{Shah, D.} \emph{et~al.}
\newblock \bibinfo{title}{{ViNT}: A foundation model for visual navigation}.
\newblock In \emph{\bibinfo{booktitle}{2023 Conference on Robot Learning (CoRL)}} (\bibinfo{year}{2023}).

\bibitem{mourikis2007multi}
\bibinfo{author}{Mourikis, A.~I.} \& \bibinfo{author}{Roumeliotis, S.~I.}
\newblock \bibinfo{title}{A multi-state constraint {Kalman} filter for vision-aided inertial navigation}.
\newblock In \emph{\bibinfo{booktitle}{Proceedings of the 2007 IEEE International Conference on Robotics and Automation}}, \bibinfo{pages}{3565--3572} (\bibinfo{year}{2007}).

\bibitem{bloesch2015robust}
\bibinfo{author}{Bloesch, M.}, \bibinfo{author}{Omari, S.}, \bibinfo{author}{Hutter, M.} \& \bibinfo{author}{Siegwart, R.}
\newblock \bibinfo{title}{Robust visual inertial odometry using a direct {EKF}-based approach}.
\newblock In \emph{\bibinfo{booktitle}{2015 IEEE/RSJ International Conference on Intelligent Robots and Systems (IROS)}}, \bibinfo{pages}{298--304} (\bibinfo{year}{2015}).

\bibitem{bloesch2017iterated}
\bibinfo{author}{Bloesch, M.}, \bibinfo{author}{Burri, M.}, \bibinfo{author}{Omari, S.}, \bibinfo{author}{Hutter, M.} \& \bibinfo{author}{Siegwart, R.}
\newblock \bibinfo{title}{Iterated extended {K}alman filter based visual-inertial odometry using direct photometric feedback}.
\newblock \emph{\bibinfo{journal}{The International Journal of Robotics Research}} \textbf{\bibinfo{volume}{36}}, \bibinfo{pages}{1053--1072} (\bibinfo{year}{2017}).

\bibitem{qin2018vins}
\bibinfo{author}{Qin, T.}, \bibinfo{author}{Li, P.} \& \bibinfo{author}{Shen, S.}
\newblock \bibinfo{title}{{VINS}-{M}ono: A robust and versatile monocular visual-inertial state estimator}.
\newblock \emph{\bibinfo{journal}{IEEE Transactions on Robotics}} \textbf{\bibinfo{volume}{34}}, \bibinfo{pages}{1004--1020} (\bibinfo{year}{2018}).

\bibitem{bisiacco2020}
\bibinfo{author}{Bisiacco, M.} \& \bibinfo{author}{Pillonetto, G.}
\newblock \bibinfo{title}{On the mathematical foundations of stable {RKHS}s}.
\newblock \emph{\bibinfo{journal}{Automatica}} \textbf{\bibinfo{volume}{118}}, \bibinfo{pages}{109038} (\bibinfo{year}{2020}).

\bibitem{Seville1968}
\bibinfo{author}{Chapman, S.}
\newblock \bibinfo{title}{Catching a baseball}.
\newblock \emph{\bibinfo{journal}{American Journal of Physics}} \textbf{\bibinfo{volume}{36}}, \bibinfo{pages}{868--870} (\bibinfo{year}{1968}).

\bibitem{Regan1997}
\bibinfo{author}{Regan, D.}
\newblock \bibinfo{title}{Visual factors in hitting and catching}.
\newblock \emph{\bibinfo{journal}{Journal of Sports Sciences}} \textbf{\bibinfo{volume}{15}}, \bibinfo{pages}{533--558} (\bibinfo{year}{1997}).

\bibitem{blondel2022efficient}
\bibinfo{author}{Blondel, M.} \emph{et~al.}
\newblock \bibinfo{title}{Efficient and modular implicit differentiation}.
\newblock \emph{\bibinfo{journal}{Advances in neural information processing systems}} \textbf{\bibinfo{volume}{35}}, \bibinfo{pages}{5230--5242} (\bibinfo{year}{2022}).

\bibitem{todorov2012mujoco}
\bibinfo{author}{Todorov, E.}, \bibinfo{author}{Erez, T.} \& \bibinfo{author}{Tassa, Y.}
\newblock \bibinfo{title}{Mujoco: A physics engine for model-based control}.
\newblock In \emph{\bibinfo{booktitle}{2012 IEEE/RSJ International Conference on Intelligent Robots and Systems}}, \bibinfo{pages}{5026--5033} (\bibinfo{year}{2012}).

\bibitem{cvxoptbook}
\bibinfo{author}{Andersen, M.~S.}, \bibinfo{author}{Dahl, J.}, \bibinfo{author}{Liu, Z.} \& \bibinfo{author}{Vandenberghe, L.}
\newblock \bibinfo{title}{Interior-point methods for large-scale cone programming}.
\newblock In \bibinfo{editor}{Sra, S.}, \bibinfo{editor}{Nowozin, S.} \& \bibinfo{editor}{Wright, S.~J.} (eds.) \emph{\bibinfo{booktitle}{Optimization for Machine Learning}}, \bibinfo{pages}{55--83} (\bibinfo{publisher}{MIT Press}, \bibinfo{year}{2012}).

\end{thebibliography}
\bibliographystyle{naturemag} 

\clearpage



\clearpage

\begin{figure}
  \centering
  \includegraphics[width=0.87\linewidth]{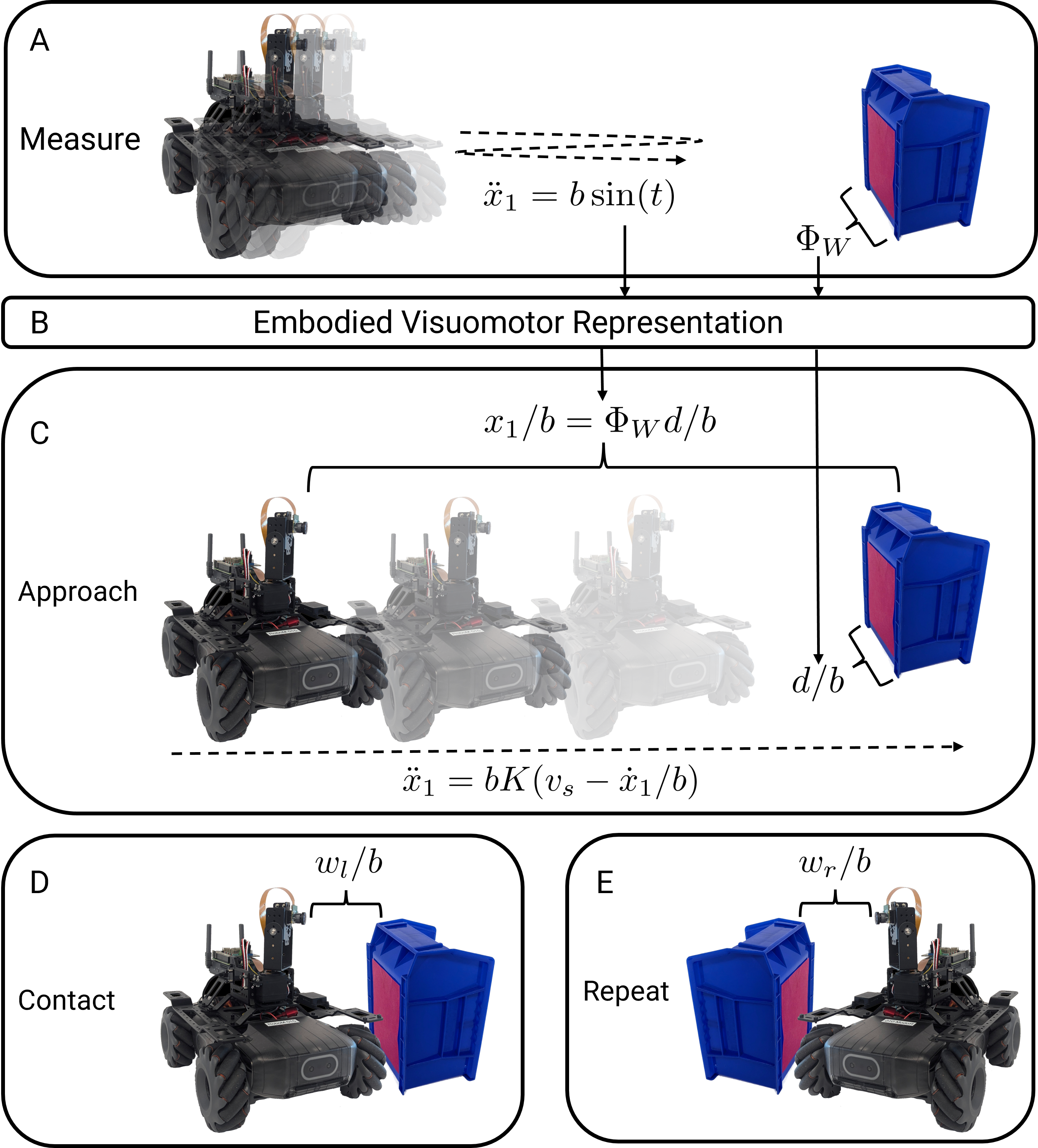}
  \caption{\textbf{Overview of the procedure for uncalibrated touching using Embodied Visual Representation.} (A): The robot oscillates by applying an open loop control input $\sin(t)$ while measuring the size of the touching target in the visual field as $\Phi_W$. (B): Embodied Visuomotor Representation uses the control input and visual information \blue{to estimate $x_1/b$ and $d/b$, which are the size of and distance to the target in the embodied unit.} (C): With the size of the target known, it is possible to approach the target at a \blue{desired safe contact speed, $v_s$,} using closed-loop control. In the closed loop, the effect of the unknown embodied gain $b$ cancels except for its interaction with the setpoint $v_s$. (D): Upon making contact with the target, the distance between the camera and the target is the size of the body $w_l$ in the embodied unit. (E): The procedure can be repeated, \blue{by approaching from different directions,} to approximate the convex hull of the robot.}
  \label{fig:touching}
\end{figure}

\begin{figure}
  \centering
  \includegraphics[width=1.0\linewidth]{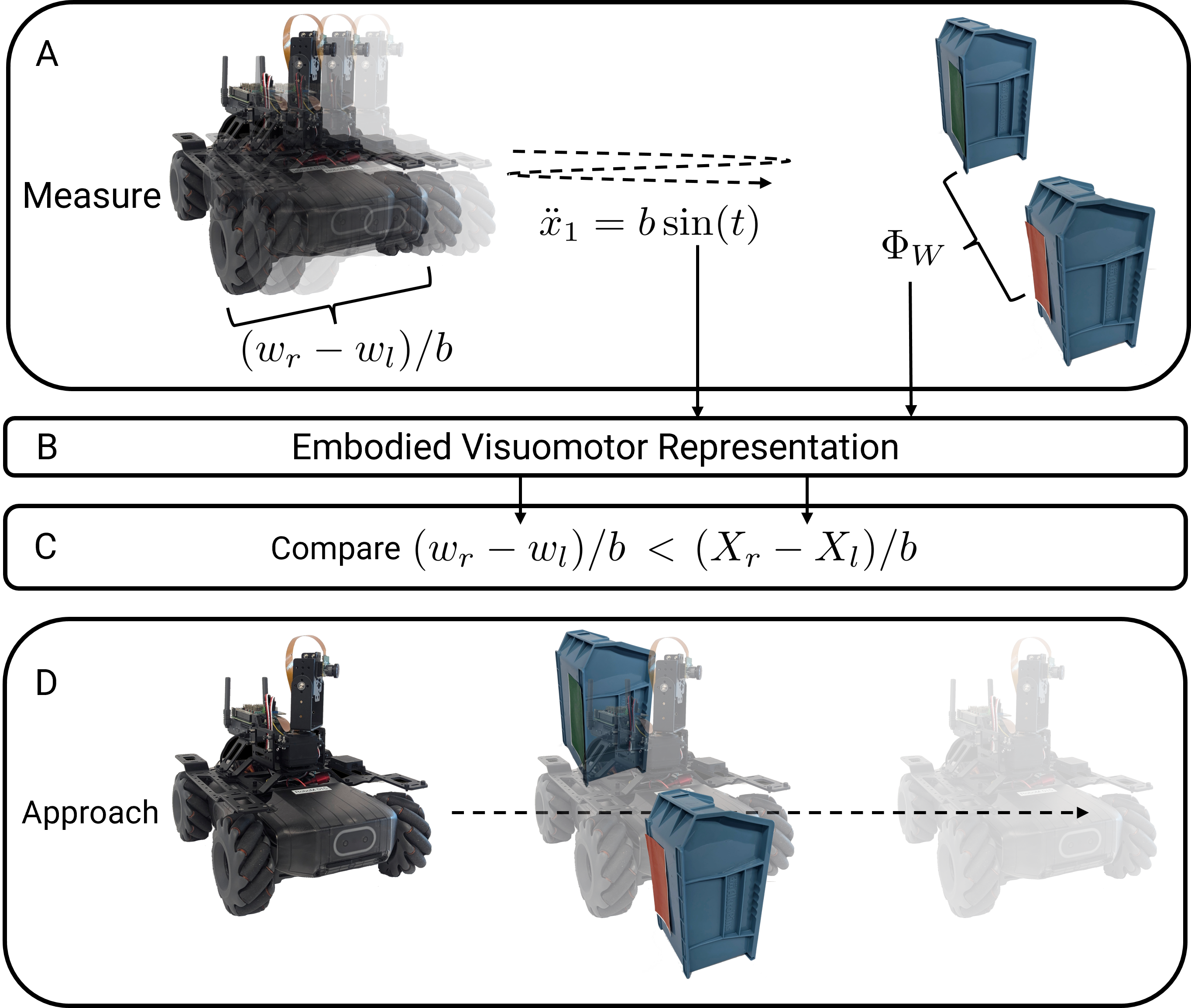}
  \caption{\textbf{Overview of the procedure for clearing obstacles using Embodied Visual Representation.} (A): The robot oscillates by applying the control input $\sin(t)$ in open loop input while \blue{using} the size of the opening between two obstacles in the visual field \blue{to estimate its distance to scale,} $\Phi_W$. (B): Embodied Visuomotor Representation uses the control input and visual information to estimate the \blue{opening size}, $(X_r-X_l) \blue{/ b}$, in the embodied units. (C): The size of the robot in the embodied unit, $(w_r - w_l)/b$, as \blue{estimated with a series of} uncalibrated \blue{touches}, is compared to the size of the opening in the embodied unit to determine if the robot can fit or clear the opening. (D) If the robot can fit, the known size of the opening can be used to determine the position of the robot\blue{,} and closed-loop control can guide the robot through the opening.}
  \label{fig:clearing}
\end{figure}

\begin{figure}
  \centering
  \includegraphics[width=1.0\linewidth]{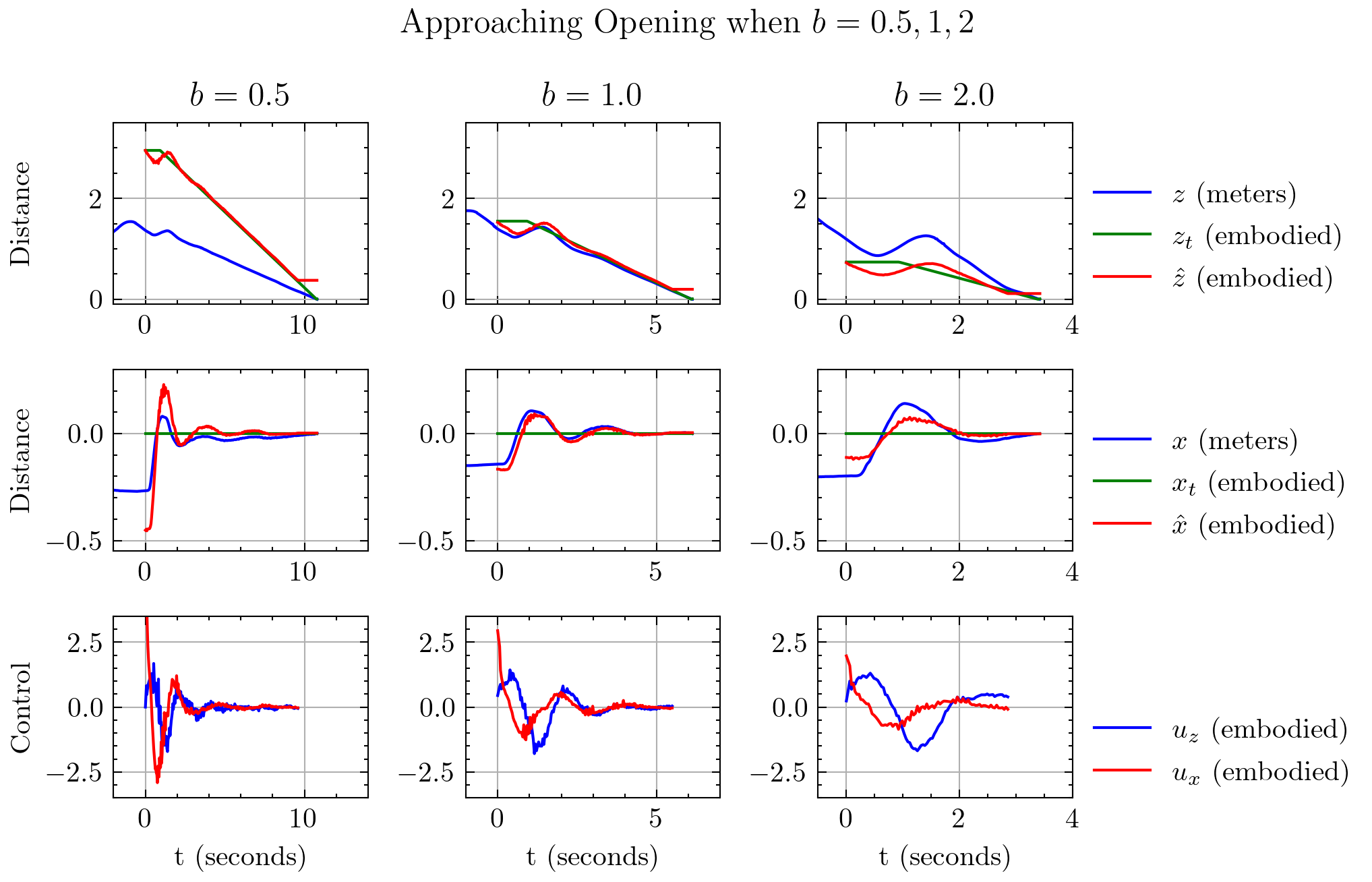}
  \caption{\blue{Experimental data and estimated signals from the opening approach phase of uncalibrated clearing for three different values of the control gain $b$. The methods take different amounts of time to reach the target because the approach speed was specified as a fixed value in the embodied unit while $b$ was varied. Note that the closed loop behavior remains stable and qualitatively similar despite the control input gain varying by a factor of 4. As shown by Equation \eqref{eqn:bcancelsdoubleint}, this is expected.}}
  \label{fig:approachopening}
\end{figure}

\begin{figure}
  \centering
  \includegraphics[width=1.0\linewidth]{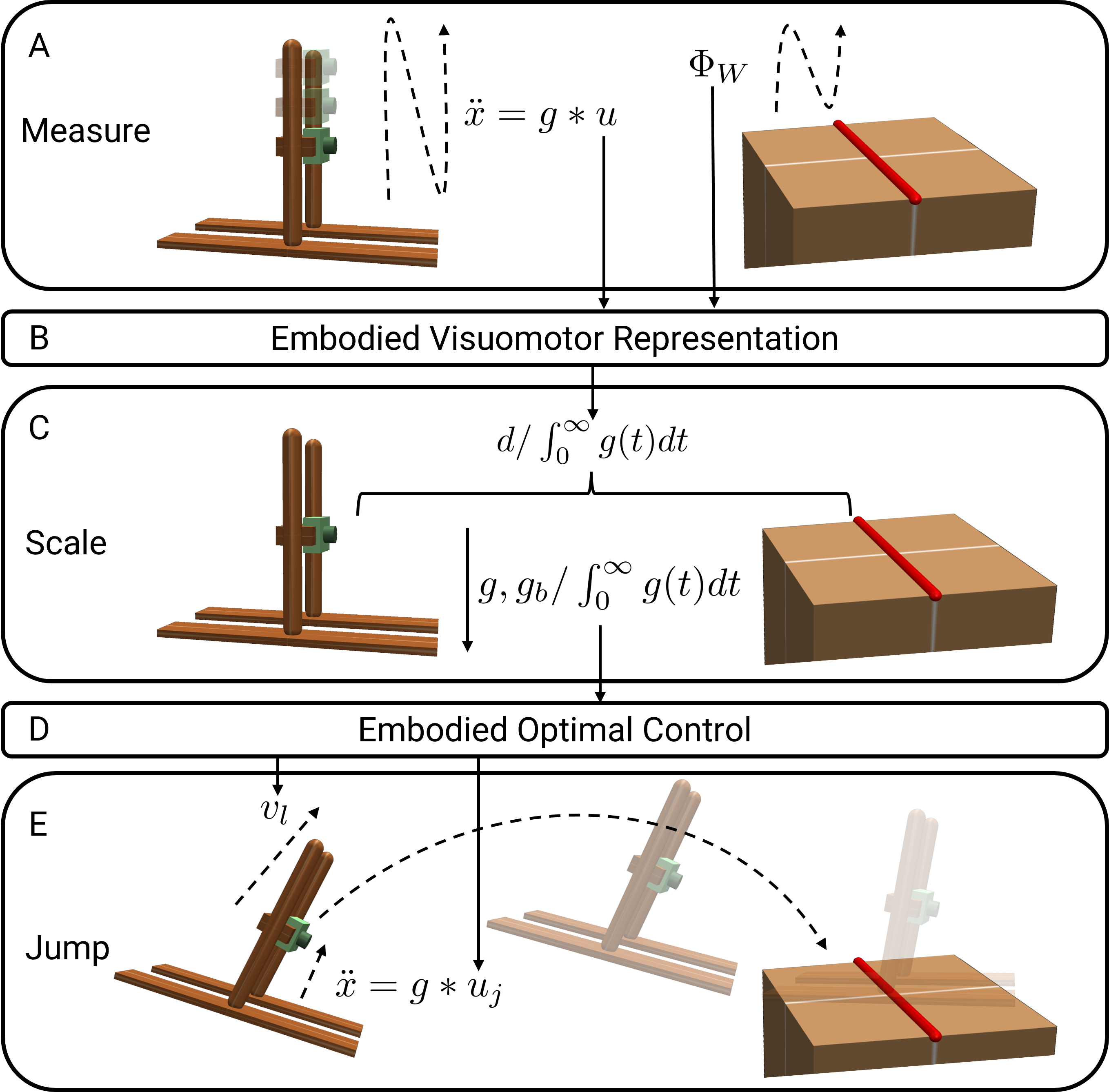}
  \caption{\textbf{Overview of the procedure for jumping a gap using Embodied Visual Representation.} (A): The robot oscillates up and down using a high gain control while measuring the control input $u$ and vertical position in the visual field $\Phi_W$ of a target line (red). The \blue{vertical} acceleration achieved \blue{$\ddot{X_2}$} results from filtering the control input by the unknown actuator dynamics $g$. (B), (C): Embodied Visuomotor Representation uses the control input and visual information to estimate the actuator dynamics $g$, distance to the jumping target $d$, and strength of gravity $g_b$ in the embodied units $\left(\int_0^{\infty}g(t) dt\right)^{-1}$. (D): An embodied optimal control problem is solved for the \blue{control input  $u_j$ of minimum total variation} that will reach the required launch velocity $v_l$ to jump the gap. (E) The jumping control input is executed in open loop after tilting the body forward.}
  \label{fig:jumping}
\end{figure}

\begin{figure}
  \centering
  \includegraphics[width=1.0\linewidth]{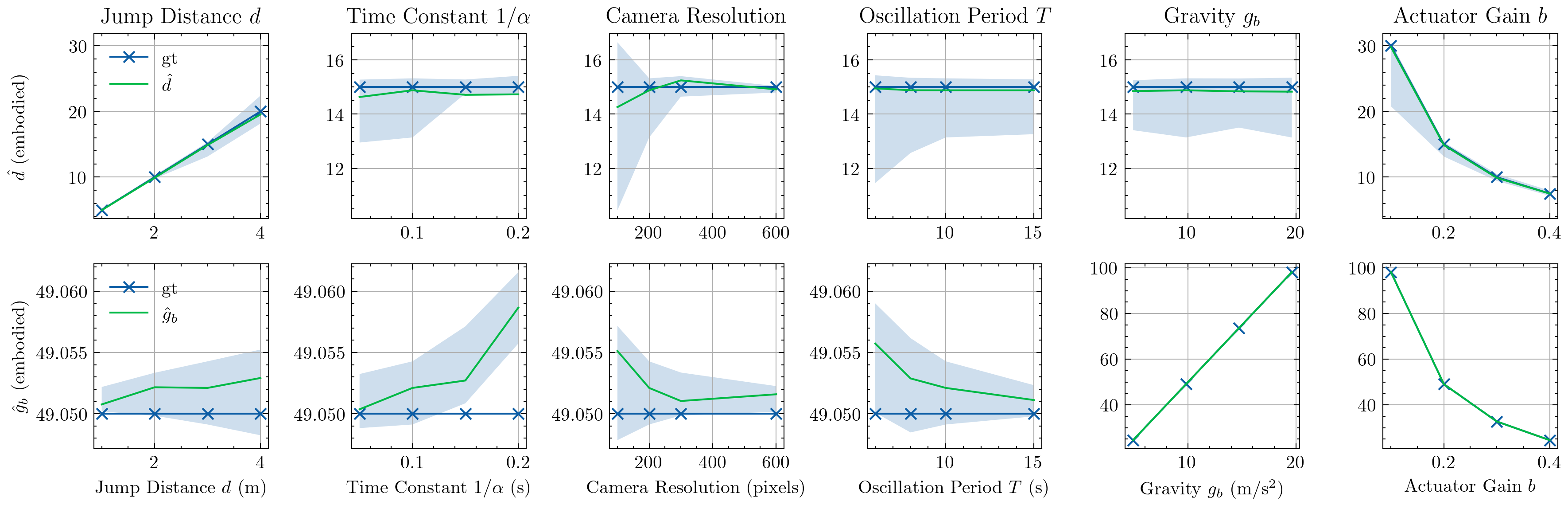}
  \caption{\blue{\textbf{Estimated jump distance and gravitational strength in embodied units versus experimental parameters.} The blue ground truth corresponds to ground truth values. Green estimates result from using an image and color thresholding to estimate $\Phi_W$. Blue shaded regions illustrate the maximum error achieved via simulated pixel quantization errors varied over 26 trials. Larger jump distances are estimated less accurately. Lowering the camera resolution below 200 pixels increases estimation error quickly. Increasing the oscillation period slightly improves results. Lower actuator gain, which results in less oscillation, decreases the accuracy of the estimated distance. While some variation in the estimate of gravitational bias can be observed, the error is typically on the order of 0.02 \%. The effects of varying the actuator time constant and gravitational strength are included for completeness when comparing to Figure \ref{fig:jumpingjumpdata}.}}
  \label{fig:jumpingestimationdata}
\end{figure}

\clearpage
\begin{table}[t]
\centering
\begin{tabular}{c|cccccc}
\hline
 & \multicolumn{6}{c}{\textbf{Initial Distance (cm)}} \\
\cline{2-7}
& 50 & 75 & 100 & 150 & 225 & 350 \\ 
\cline{2-7}
\textbf{Control Gain} & \multicolumn{6}{c}{\textbf{Target Width Error (\%)}} \\
\hline
$b=0.5$ & 1.00 & 0.61 & 1.37 & 4.62 & 7.42 & 20.80 \\ 
$b=1.0$ &  -   & 0.66 & 1.17 & 3.05 & 5.91 & 3.61 \\
$b=2.0$ &  -   &  -   &  -   & 14.87 & 7.00 & 6.53 \\
\hline
&\multicolumn{6}{c}{\textbf{Body Width Error (\%)}} \\
\hline
$b=0.5$ & 4.51 & 6.08 & 5.33 & 2.33 & 7.19 & 21.28 \\
$b=1.0$ &  -   & 6.00 & 6.90 & 3.16 & 11.52 & 6.08 \\
$b=2.0$ &  -   &  -   &  -   & 12.30 & 2.01 & 3.66 \\
\hline
\end{tabular}
\caption{\blue{Mean absolute error in percent of the estimated target width (15 cm) and the observable robot body's width (19.1 cm) as the initial distance from the target is increased. Each configuration was tested over 5 trials. The true robot body's width is 25.4 cm. However, the observable width is reduced by the pan-tilt camera's turning radius. At 350 cm, the target occupied 6 pixels in the field of view.}}
\label{table:initialdistvar}
\end{table}

\begin{table}[t]
\centering
\begin{tabular}{c|ccc|ccc}
\hline
 & \multicolumn{3}{c|}{\textbf{Target Width (cm)}} & \multicolumn{3}{c}{\textbf{Opening Width (cm)}} \\
\cline{2-7}
& 3.75 & 7.5 & 15 & 12.5 & 25 & 37.5 \\ 
\cline{2-7}
\textbf{Control Gain} & \multicolumn{3}{c|}{\textbf{Target Width Error (\%)}} & \multicolumn{3}{c}{\textbf{Opening Width Error (\%)}}  \\
\hline
$b=0.5$ & 13.00 & 11.45 &  5.45 &  8.22 &  2.21 &  3.16 \\ 
$b=1.0$ &  5.25 & 6.91  &  1.10 &  7.03 &  2.91 &  2.20 \\
$b=2.0$ & 14.79 & 36.85 & 13.15 & 16.99 & 14.96 & 17.52 \\
\hline
& \multicolumn{3}{c|}{\textbf{Body Width Error (\%)}} & & & \\
\hline
$b=0.5$ & 19.88 &  4.74 &  4.14 &  &  &  \\
$b=1.0$ & 21.39 &  8.98 &  5.69 &  &  &  \\
$b=2.0$ & 12.57 & 20.66 & 10.07 &  &  &  \\
\hline
\end{tabular}
\caption{\blue{Mean absolute error in percent of the estimated touch target or opening width and observable robot body's width (19.1 cm) as the target or opening width is varied. Each configuration was tested over 5 trials. The true robot body's width is 25.4 cm. However, the observable width is reduced by the pan-tilt camera's turning radius. The initial distance was set to 150 cm. Only the touch target trials are associated with an estimated body size.}}
\label{table:targetvar}
\end{table}

\begin{table*}
\small
  \centering
  \begin{tabular}{@{}lp{6.5cm}p{7.0cm}}
    \\[0.03cm]
 \textbf{Task} & \textbf{External Scale} & \textbf{Embodied Visuomotor Representation (EVR)} \\
    \midrule
    Touching &
    \parbox{6.5cm}{State estimated with calibrated sensors in external scale \\ \\
                 Control stabilizes state via conversion to embodied signals \\ \\
                 Stability depends on \blue{prior} knowledge of conversion factor\\} &
    \parbox{6.5cm}{Embodied state estimated with EVR \\ \\ \\
                 Control stabilizes embodied state with embodied signals \\ \\
                 Stability is invariant to units \\ \\}\\
    \midrule
    Clearing &
    \parbox{6.5cm}{Robot's physical size defined in external scale \\ \\
                 State estimated with calibrated sensors in external scale \\  \\
                 The clearing algorithm uses a world in an external scale and clears objects via conversion to embodied signals \\ \\
                 Stability depends on conversion and a pre-determined physical model \\} &
    \parbox{6.5cm}{Embodied physical size estimated with touch \\ \\
                   Embodied state estimated with EVR \\ \\ \\
                   The clearing algorithm uses an embodied world that is consistent with embodied signals\\ \\
                   Stability is invariant to units and physical dimensions \\} \\
    \midrule
    Jumping &
    \parbox{6.5cm}{
    Jumping distance estimated using calibrated sensors \\ \\
    Launch velocity estimated in external scale \\ \\
    Control achieves launch velocity via conversion to embodied signals  \\ \\
    Success dependent on scale and pre-determined controller} &
    \parbox{6.5cm}{Jumping distance and actuator dynamics estimated with EVR \\  \\
                  Launch control signal estimated via embodied representation\\ \\
                  Launch control signal applied open loop \\ \\
                  Stability invariant to units and controller} \\
    \bottomrule
  \end{tabular}
  \caption{Robot behaviors that can be accomplished with external scale and traditional methods, but can also be accomplished with Embodied Visuomotor Representation (EVR). Due to its use of embodied scale, Embodied Visuomotor Representation theoretically guarantees success and stability without calibrated sensors, pre-determined physical models, and low-level controllers that are tuned prior to deployment.}
  \label{tab:embodimentmodes}
\end{table*}

\clearpage
\section*{Supplementary Material}
\setcounter{equation}{0}

\blue{\subsection*{Supplementary Movie 1}
Supplementary Movie 1 demonstrates uncalibrated clearing and touching. The robot does not know its size, strength, or the size of anything in the world and must determine if it fits through an opening of unknown size. To do so, it follows the procedure illustrated by Figure \ref{fig:touching} and Figure \ref{fig:clearing}.}
\blue{\subsection*{Supplementary Movie 2}
Supplementary Movie 2 demonstrates uncalibrated jumping. The robot does not know its size, strength, leg actuator time constant, or the strength of gravity and must jump a gap of unknown size. To do so, it follows the procedure illustrated by Figure \ref{fig:jumping}.}
\blue{\subsection*{Proofs}
\subsubsection*{Uniqueness of solution of \eqref{eqn:embodieddoubleintest}}
Equation \eqref{eqn:embodieddoubleintest} has a unique solution if and only if acceleration is non-zero for some time in the interval. \eqref{eqn:embodieddoubleintest} is reproduced below for clarity:}

\blue{\begin{equation*}
\begin{split}
\min_{\frac{\left[d, x(t_0)\right]}{b}} \int_{t_0}^{t_0+T} \left(
\Phi_W(t, t_0) \frac{d}{b} - \frac{x_1(t_0)}{b} - (t - t_0)\frac{x_2(t_0)}{b} - \int_{t_0}^t \int_{t_0}^\sigma u(\sigma_2) d\sigma_2 d\sigma \right)^2 dt
\end{split}
\end{equation*}}
\blue{\textbf{Proof}
The problem is a linear least squares problem. Thus it has a unique solution if and only if the signals $\Phi_W(t, t_0)$, $1$, and $t-t_0$, are linearly independent over the interval $[t, t_0+T]$. Immediately, we can see that the signals $1$ and $t-t_0$ are independent over any interval of non-zero size. It remains to show under what conditions they are independent of $\Phi_W(t, t_0)$. Consider that, by definition, $\Phi_W$ and $u$ are related by}

\blue{
\begin{equation}
\Phi_W(t, t_0)d = x_1(t) = x_1(t_0) - (t - t_0) x_2(t_0) - b \int_{t_0}^t \int_{t_0}^\sigma u(\sigma_2) d\sigma_2 d\sigma.
\end{equation}
}

\blue{
Thus, $\Phi_W$ is linearly independent from $1$ and $t-t_0$ if and only if the double integral of $u$ is independent from $1$ and $t-t_0$. Suppose they are linearly dependent. Then there exists $\alpha_1, \alpha_2$ such that for all $t \in [t_0, t_0+T]$
\begin{equation}\label{eqn:doubleintli}
\alpha_1 + (t - t_0) \alpha_2 = \int_{t_0}^t \int_{t_0}^\sigma u(\sigma_2) d\sigma_2 d\sigma.
\end{equation}
}

\blue{
Taking the derivative of both sides twice implies $u(t) = 0$ for all $t$ in the interval. Similarly, if $u(t) = 0$ for all $t$ then the signals are linearly dependent. Thus, \eqref{eqn:embodieddoubleintest} has a unique solution, that is, $\Phi_W$ is linearly independent from $1$ and $t-t_0$, if and only if $u(t) \neq 0$ for some time in the interval $[t_0, t_0+T]$.
}

\blue{
\subsubsection*{Uniqueness of solution of \eqref{eqn:embodieddoubleintestunbiased}}
Equation \eqref{eqn:embodieddoubleintestunbiased} has a unique solution if and only if acceleration is non-zero for some time in the interval. It is reproduced below for clarity:
}

\blue{
\begin{equation*}
\begin{split}
\min_{\frac{\left[b, x(t_0)\right]}{d}} \int_{t_0}^{t_0+T} \left(
\Phi_W(t, t_0) - \frac{x_{\blue{1}}(t_0)}{d} - (t - t_0)\frac{\blue{x_2}(t_0)}{d} - \frac{b}{d}\int_{t_0}^t \int_{t_0}^\sigma u(\sigma_2) d\sigma_2 d\sigma \right)^2 dt.
\end{split}
\end{equation*}
}

\blue{
\textbf{Proof:}
The problem is a linear least squares problem and thus has a solution if only if the double integral of $u$ is linearly independent of 1 and $t-t_0$. The proof can proceeds identically to the previous section starting from equation \eqref{eqn:doubleintli}.
}

\blue{
\subsubsection*{\eqref{eqn:embodieddoubleintestunbiased} is an unbiased estimator}
}

\blue{
\textbf{Proof:}
Equation \eqref{eqn:embodieddoubleintestunbiased} is a linear least squares problem. Thus, its solution is linear in $\Phi_W$. In particular, the solution takes the form
}

\blue{
\begin{equation}
\frac{\left[b, x(t_0)\right]}{d}^* = \langle J, J\rangle ^{-1} \langle J, \Phi_W \rangle .
\end{equation}
Here $J = [1, t-t_0, \iint u]$ is a vector of signals and $\langle \cdot, \cdot\rangle $ denotes the continuous time inner product evaluated over the interval $[t_0, t_0+T]$. Consequently, if $\Phi_W$ is corrupted with zero mean noise $\mathcal{N}$, then expectation distributes and the estimate is unbiased. That is,
}

\blue{
\begin{equation}
\begin{split}
E\left[\frac{\left[b, x(t_0)\right]}{d}^*\right] &= E\left[\langle J, J\rangle ^{-1} \langle J, \Phi_W + \mathcal{N}\rangle\right] \\
&= \langle J, J\rangle ^{-1} \langle J, E\left[\Phi_W + \mathcal{N}\right]\rangle \\
&= \langle J, J\rangle ^{-1} \langle J, \Phi_W + E\left[\mathcal{N}\right]\rangle \\
&= \langle J, J\rangle ^{-1} \langle J, \Phi_W\rangle. \\
\end{split}
\end{equation}
Thus, the estimator is unbiased.
}

\blue{
Further, the population average of $E[\mathcal{N}]$ will converge to zero as the length of the considered interval $T$ increases. Thus, longer estimation intervals will result in better results. Additionally, the dependence of the solution on $\mathcal{N}$ can be decreased by increasing the magnitude of $J$ and $\Phi_W$. A system can accomplish this by making motions of greater amplitude.
}

\blue{
\subsubsection*{The error of the estimate due to \eqref{eqn:embodieddoubleintestunbiased} decreases with the inverse of camera resolution}
}

\blue{
Suppose $\Phi_W$ is quantized due to its measurement by a camera of finite resolution. That is $\hat{\Phi}_W = \mathrm{round}(\Phi_W / \Delta) * \Delta$ where $\Delta \propto 1 / \mathrm{res}$ and $\mathrm{res}$ is the resolution of the camera. Then the error in solving \eqref{eqn:embodieddoubleintestunbiased}'s deccreases with the inverse of the camera's resolution.
}

\blue{
\textbf{Proof:}
Recall from the previous section that
\begin{equation}
x^* = \langle J, J\rangle ^{-1} \langle J, \Phi_W \rangle,
\end{equation}
where we let $x^*$ be the vector of solution parameters for notational convenience. By linearity of this solution, we have that the norm of the perturbed solution is
\begin{equation}
\|dx^*\| = \left\|\langle J, J\rangle ^{-1} \langle J, \left(\Phi_W - \hat{\Phi}_W\right) \rangle\right\| \leq \left\|\langle J, J\rangle ^{-1} \langle J, \cdot\rangle\right\| \left\| \Delta \right\|
\end{equation}
}

\blue{
Thus, we can expect errors in parameter estimates to decrease like $\Delta = 1/\mathrm{res}$. This is confirmed by Figure \ref{fig:jumpingestimationdata}. While Figure \ref{fig:jumpingestimationdata} is produced by solving Equation \eqref{eqn:generalizedphiconst}, the same argument holds for bounding the error in the estimates due to finite resolution.
}

\blue{
\subsubsection*{Uniqueness of solution of \eqref{eqn:slidingadaptiveproblem}}
The estimation of the parameters of linear systems is possible in general if and only if the input $u$ is sufficiently exciting. This is a well known result that applies to all linear systems. An overview of the traditional approach is given in Chapter 2 of \cite{astrom2008adaptive} which briefly develops the general case and considers in detail the use of a single input to predict a single output using an FIR filter.
}

\blue{
Here we develop a specific excitation condition for the case of \eqref{eqn:slidingadaptiveproblem}. That is for the the case of multiple inputs and double integrated linear actuator dynamics. Further, because an impulse response is infinite-dimensional, we require that it be represented with a finite-dimensional basis. Using a finite dimensional basis is required for implementation on digital hardware. Further, while textbooks such as \cite{astrom2008adaptive} focus on identifying finite impulse response (FIR) filters, this simply corresponds to the implicit assumption of a basis of delayed delta-diracs. Thus, identifying FIR filter parameters is a special case of the following result.
}

\blue{
\eqref{eqn:slidingadaptiveproblem} is reproduced below for convenience:
\begin{equation*}
\begin{split}
    \min_{\frac{[X(t_0), \dot{X}(t_0), G]}{d}} 
    \int_{t_0}^{t_0+T} \Bigg[&\Phi(t, t_0) -\frac{X(t_0)}{d} -(t-t_0) \frac{\dot{X}(t_0)}{d} 
    - \left(\frac{G}{d} * \int_{t_0}^{\boldsymbol{\cdot}} \int_{t_0}^{\sigma} u (\sigma) d\sigma_2 d\sigma  d\sigma\right)(t)\Bigg]^2 dt,
\end{split}
\end{equation*}
}

\blue{
Where $G$ is a matrix value signal whose $i$,$j$'th entry is to be convolved with the $j$'th input to get its effect on the $i$'th output. Let the $i,j$'th entry be denoted by $g^{ij}$. Suppose each $g^{ij}$ is parameterized by a set of basis functions. That is $g^{ij} = \sum_k c^{ij}_k f_k = {c^{ij}}^T f$ where $f_k$ is a basis function and $f=[f_1, f_2, \ldots, f_n]$ is the vector of them.
}

\blue{
The resulting finite-dimensional problem is:
\begin{equation}\label{eqn:slidingadaptiveproblemfinite}
\begin{split}
    \min_{\frac{[X(t_0), \dot{X}(t_0)]}{d}, c}
    \int_{t_0}^{t_0+T} \Bigg[&\Phi(t, t_0) -\frac{X(t_0)}{d} -(t-t_0) \frac{\dot{X}(t_0)}{d} 
    - \left(
    \underbrace{\begin{bmatrix}
        {c^{ij}}^T f
    \end{bmatrix}}_{G/d}
    * \int_{t_0}^{t} \int_{t_0}^{\sigma} u (\sigma) d\sigma_2 d\sigma  d\sigma\right)(t)\Bigg]^2 dt,
\end{split}
\end{equation}
}

\blue{
In what follows, we show that the problem has a unique solution if and only if an excitation condition is satisfied. Additionally, it is necessary that the set of inputs $\{u_1, \ldots, u_M\}$ and basis functions $\{f_1, \ldots, f_N\}$ be respectively linearly independent.
}

\blue{
\textbf{Proof:}
The problem consists of three separate linear least squares problems, one for each element of $\Phi$. It then immediately follows that these problems have a solution if and only if the sets of signals $S \coloneqq \{1, t-t_0\} \cup \{f_k * \int_{t_0}^{\cdot} \int_{t_0}^{\sigma} u_j(\sigma_2) d\sigma_2 d\sigma \, | \, \forall k, j\}$ are linearly independent. Note that this one condition is sufficient and necessary for all three problems.
}

\blue{
Suppose that the set of signals $S$ is linearly dependent. Then there exists scalars  $\alpha_1$, $\alpha_2$ and a vector $\beta$ such that for all $t \in [t_0, t_0+T]$
\begin{equation}\label{eqn:implindep}
\alpha_1 + (t - t_0) \alpha_2 = \sum_j \sum_k \beta_{jk} \left(f_k * \int_{t_0}^{\cdot} \int_{t_0}^\sigma u_j d\sigma_2 d\sigma\right)(t).
\end{equation}
}

\blue{
We can move the convolutions with $f_k$ inside the double integral and take the derivative twice, revealing that:
\begin{equation}
 \sum_j \sum_k \beta_{jk} \left(f_k * u_j \right) = 0
\end{equation}
}

\blue{
Therefore, $S$ is linearly independent if and only if $\exists t$ such that $\sum_j \sum_k \beta_{jk} \left(f_k * u_j \right)(t) \neq 0$. Consequently, it is necessary that the elements of $f$ and $u$ be linearly independent since
\begin{equation}
 \sum_j \sum_k \beta_{jk} \left(f_k * u_j \right) = \sum_j \left(\sum_k \beta_{jk} f_k\right) * u_j = \sum_k f_k * \left(\sum_j \beta_{jk} u_j\right),
\end{equation}
}

\blue{
and so linear dependence of the elements of either $u$ or $f$ implies the existence of non-zero $\beta$ such that the right hand side of \eqref{eqn:implindep} is zero. Then linear dependence of $S$ holds with $\alpha_1, \alpha_2 = 0$.
}

\blue{
Finally, we realize the excitation condition necessary for a unique solution. Define 
\begin{equation}
\phi \coloneqq \mathrm{vec}\left\{
\begin{bmatrix}
(f_1 * u_1) & (f_2 * u_1) & \ldots & (f_N * u_1)\\
(f_1 * u_2) & (f_2 * u_2) & \ldots & (f_N * u_2)\\
\vdots & \vdots & \ddots & \vdots\\
(f_1 * u_M) & (f_2 * u_M) & \ldots & (f_N * u_M)\\
\end{bmatrix}
\right\}
=
\mathrm{vec}\{u * f^T\}.
\end{equation}
}

\blue{
Where $\mathrm{vec}$ transforms a matrix to a vector and convolution is performed elementwise.
}

\blue{
Then, if linear independence holds, $\forall \beta \neq 0$ we have that $\phi^T \beta \neq 0$. Squaring and integrating over time results in an excitation condition. That is, \eqref{eqn:slidingadaptiveproblemfinite} has a unique solution if and only if the following excitation condition is satisfied:
\begin{equation}
\int_t^{t+T} \phi(\sigma) \phi^T(\sigma) d\sigma > 0 .
\end{equation}
}

\blue{
In practice, it is well known that inputs that excite a wide range of frequencies result in good persistence of excitation \cite{astrom2008adaptive}.
}

\blue{
\subsection*{Choice of safe contact speed $v_s$}
}

\blue{
Determining the safe contact speed during uncalibrated touching requires some assumptions by the designer or a higher level cognitive process within the robot. In general, and as with biological systems, there is no way to determine at what speed a system can make its first contact with the world without breaking. Below, we give two examples of how to set $v_s$ for an embodied system.
}

\blue{
\subsubsection*{Multiples of the touch target's size}
Suppose the robot is to touch an object that is 0.5 meters wide, and the width of the touch target is the characteristic scale of vision $d$. Further, suppose the robot's designer wants the robot to come into contact with the target while traveling at $0.2$ meters per second. Then, since the safe contact speed is equivalent to $0.2/0.5 = 0.4$ multiples of $d$ per second, the safe contact speed should be set as $v_s = 0.4d$ within the robot's software. Subsequently, the robot can convert this setpoint to embodied units using its own estimate of $d$.
}

\blue{
\subsubsection*{Multiples of maximum force}
It is also possible to determine $v_s$ by assuming a simple contact model, constant deceleration over some time, and requiring that the magnitude of the acceleration be equal to some multiple of the maximum force the robot can exert on itself. Let this multiple be $\alpha$. Then, suppose the maximum acceleration achievable by the system's own actuators is $u^{max}$, and the expected deceleration period is $T>0$. In that case, the average force experienced on contact is $v_s / T$, which should be less than $\alpha u^{max}$. Thus, the safe speed can be chosen to satisfy $v_s <= \alpha T u^{max}$ where $\alpha$ and $T$ can be intuitively chosen by the robot's designer.
}

\end{document}